\documentclass{article}
\usepackage{ijcai23}

\usepackage{times}
\usepackage{soul}
\usepackage{url}
\usepackage[hidelinks]{hyperref}
\usepackage[utf8]{inputenc}
\usepackage[small]{caption}
\usepackage{graphicx}
\usepackage{amsmath}
\usepackage{amsthm}
\usepackage{booktabs}
\usepackage{algorithm}
\usepackage{algorithmic}
\usepackage[switch]{lineno}
\usepackage{comment}

\usepackage{subfiles}
\usepackage[capitalize,noabbrev]{cleveref}
\crefformat{equation}{(#2#1#3)}
\crefrangeformat{equation}{#3(#1)#4 to #5(#2)#6}
\usepackage[dvipsnames]{xcolor}
\usepackage{amsmath}
\usepackage{amssymb}
\usepackage{enumitem}
\usepackage{caption}
\usepackage{subcaption}

\newtheorem{definition}{Definition}

\DeclareMathOperator*{\argmax}{arg\,max}
\DeclareMathOperator*{\KL}{\operatorname{KL}}
\DeclareMathOperator*{\dis}{\operatorname{dis}}

\newcommand{\citet}[1]{\citeauthor{#1}~\shortcite{#1}}
\newcommand{\citep}{\cite}

\title{Revisiting Estimation Bias in Policy Gradients for Deep Reinforcement Learning}

\author{
    Haoxuan Pan$^{1,2}$
    \and
    Deheng Ye$^{2}$\and 
    Xiaoming Duan$^1$\and\\
    Qiang Fu$^2$\and
    Wei Yang$^2$\and
    Jianping He$^1$ \and
    Mingfei Sun$^3$
    \affiliations
    $^1$ Shanghai Jiao Tong University \quad
    $^2$ Tencent Inc. \quad
    $^3$ The University of Manchester
    \emails
    \{panhaoxuan, xiaoming, jphe\}@sjtu.edu.cn,
\{dericye,leonfu,willyang\}@tencent.com,    
mingfei.sun@manchester.ac.uk
}

\begin{document}

\maketitle

\begin{abstract} 
We revisit the estimation bias in policy gradients for the discounted episodic Markov Decision Process (MDP) from Deep Reinforcement Learning (DRL) perspective. 
The objective is formulated theoretically as the expected returns discounted over the time horizon. 
One of the major policy gradient biases is the state distribution shift: 
the state distribution used to estimate the gradients differs from the theoretical formulation 
in that it does not take into account the discount factor. 
Existing discussion of the influence of this bias was limited to the tabular and softmax cases in the literature. 
Therefore, in this paper, we extend it to the DRL setting where the policy is parameterized and demonstrate how this bias can lead to suboptimal policies theoretically. 
We then discuss why the empirically inaccurate implementations with shifted state distribution can still be effective. 
We show that, despite such state distribution shift, 
the policy gradient estimation bias can be reduced in the following three ways: 
1) a small learning rate; 2) an adaptive-learning-rate-based optimizer; and 3) KL regularization. 
Specifically, we show that a smaller learning rate, or, an adaptive learning rate, 
such as that used by Adam and RSMProp optimizers, makes the policy optimization robust to the bias. 
We further draw connections between optimizers and the optimization regularization to show that 
both the KL and the reverse KL regularization can significantly rectify this bias. 
Moreover, we provide extensive experiments on continuous control tasks to support our analysis. 
Our paper sheds light on how successful PG algorithms optimize policies in the DRL setting, 
and contributes insights into the practical issues in DRL. 
\end{abstract}

\section{Introduction}

We consider the policy gradients (PG) for the episodic Markov Decision Process (MDP) setting 
in which the objective is to maximize the expected returns discounted over the time horizon.
In this setting, the PG derived in the seminal work~\cite{sutton1999policy} 
is defined with respect to a discounted state distribution, 
i.e., the state probability in the Markov chain induced by the underlying policy is discounted over the time horizon.
In practice, however, 
many empirically successful algorithms, e.g., Trust Region Policy Optimization (TRPO)~\cite{schulman2015trust}, 
Proximal Policy Optimization~\cite{schulman2017proximal},
do not strictly follow this form of state distribution\cite{nota2020policy}. 
Instead, they often drop the discount factor in the state distribution, i.e., undiscounted state distribution, 
leading to a shifted state distribution, 
and the resulting PG estimate is thus biased~\cite{nota2020policy}.

On the other hand, it still remains largely unclear how the biased PG will influence the policy optimization in neural network (NN) parameterization, i.e., Deep Reinforcement Learning (DRL). 
This bias was first observed in~\cite{thomas2014bias} which shows
that many PG algorithms are implemented with respect to the undiscounted state distribution.
\cite{nota2020policy} further proved that the widely used undiscounted ``policy gradient'' is not a gradient of any objective, 
and the learned policy can be highly sub-optimal under some circumstances.
\cite{wu2022understanding} further proved that, 
in the discounted MDP setting, 
the bias incurred by the state distribution shift is non-negligible.
Though some recent studies have discussed the convergence issue of PG with state distribution shift.
For example, \cite{agarwal2021theory} discussed the influence of the general distribution shift on the convergence under softmax tabular policy parameterizations.
\cite{laroche2021dr} proposed a sufficient and necessary condition for the convergence of the PG under direct and softmax parameterization.
Their conclusion may not generalize to the deep RL situation where the policy is parameterized in neural networks.

In this paper, we first discuss the \textit{state alias} phenomenon that could happen in DRL 
and will potentially result in suboptimal training for the biased PG.
We investigate the situation when this phenomenon is likely to happen, 
and explain how the state alias phenomenon is connected to the previous analysis on the PG estimate bias~\cite{nota2020policy}.
We particularly show that when this phenomenon happens
the policy optimization with a biased PG leads to a highly sub-optimal policy theoretically.

Furthermore, we discover three common policy optimization techniques that could play a significant role in reducing the bias in the PG estimate.
Specifically, we find that a model trained with a relatively small learning rate is less likely 
to be influenced by the PG bias.
Second, we find that the widely used adaptive-learning-rate-based optimizers, 
such as RMSProp~\citep{Tieleman2012RMSProp} and Adam~\cite{kingma2014adam}, 
help fix the bias. 
These optimizers adjust per-parameter learning rates based on statistics of the second moment of the gradients, 
which can also be viewed as estimating an empirical Fisher Information Matrix (FIM)~\cite{amari2012differential} 
-- leveraging the curvature of the loss surface to reduce the side-effects of the PG bias.
Third, inspired by the connection between the FIM-based method and KL regularization, 
we study the influence of KL regularization on bias fixing.
We find both KL divergence and the reverse KL divergence contribute to bias-fixing.
We show the regularized objective function gives a large penalty on the suboptimal areas.

At last, we conduct extensive experiments using Mujoco~\cite{brockman2016openai} to support our analysis. 
Specifically, we show that the vanilla biased PG performs much worse than the unbiased one, 
while the biased PG with the aforementioned techniques can maintain the performance.
We also provide a bias spread analysis to demonstrate how different methods could be influenced by the bias, during each training epoch.
Furthermore, we show through empirical experiments that those techniques can also mitigate the influence of PG biases in more general cases.

To sum up, our contributions are:
\begin{itemize}
    \item To the best of our knowledge, we are the first to theoretically investigate, in the DRL setting, the influence of the biased PG with state distribution shift. We introduce the state alias phenomenon under which the biased PG converges to a suboptimal policy with a concrete example.
    \item We investigate why the widely used biased PG can still work.
    We understand it by digging out three techniques that unexpectedly mitigate the influence of the bias: learning rate, adaptive optimizers, and regularizations.
    \item We provide extensive experiments on continuous control tasks to support our findings. 
\end{itemize}

\section{Related Work}
\paragraph{Bias analysis of PGs.}
1) We first discuss related work on investigating the causes and impact of PG bias.
\cite{thomas2014bias} first drew attention to the discrepancy between the theory and the practical implementation
that many PG implementations did not strictly follow the theory in \cite{sutton1999policy}.
\cite{nota2020policy} proved that in the worst case the undiscounted PG is not a gradient of any objective.
\cite{wu2022understanding} re-interpreted this discrepancy through a unified framework 
and suggested that a large (close to $1.0$) discount factor and experience replay help fix the bias.
2) Follow-up studies consider how to rectify the PG bias issue.
DDPG~\cite{silver2014deterministic} and its variant~\cite{fujimoto2018addressing,wu2020reducing} are examples of per-step correction, 
aiming for an accurate estimate of the PG.
\cite{agarwal2021theory} discussed how state distribution shift affects the convergence and performance of PG methods.
A sufficient and necessary condition for policy optimization to converge to the optimal policy in tabular or softmax policy parameterizations is provided by~\cite{laroche2021dr}.
Their works relax the condition of the state distribution of~\cite{sutton1999policy}.
However, the influence of the biased PG with undiscounted state distribution when the policy is parameterized with neural networks in DRL is unclear.  

\paragraph{Performance of optimizers in RL.}
Optimizer designs and selections have long been studied in the deep learning community, 
and the classic methods include AdaGrad~\cite{duchi2011adaptive}, RMSProp~\cite{Tieleman2012RMSProp} and Adam~\cite{kingma2014adam}. 
These methods estimate the first and second moments of the gradient to adaptively adjust the parameter-wise learning rate.
Plenty of studies have analyzed the convergence and performance of these optimizers in supervised learning~\cite{reddi2019convergence,zou2019sufficienta}.
However, very few studies~\cite{stooke2019rlpyt,engstrom2019implementation} discuss how an optimizer can influence the performance of DRL agents.

\paragraph{Regularized policy optimization.}
KL divergence (relative entropy) is one of the commonly used regularizers in policy optimization.
TRPO~\cite{schulman2015trust} and its variants~\cite{nachum2018trustpcl,kuba2022trust,jacob2022modeling} adopted KL divergence to constrain the policy update in each epoch.
However, constrained optimization is computationally intensive, 
thus some methods avoid solving the exact constrained optimization through relaxing the constraint~\cite{schulman2017proximal,cobbe2021phasic} or analytically solving the constrained optimization with expectation-maximization~\cite{abdolmaleki2018maximum,abdolmaleki2018relative,hessel2021muesli}.
There are also methods investigating the influence of regularization during the policy optimization process.
For example, \cite{vieillard2020leverage} revealed KL regularization implicitly averages Q-values.
The convergence rate of TRPO-based methods was studied in~\cite{shani2020adaptive,zhan2021policy}.
\cite{lazic2021optimization} suggested that using KL divergence as a regularizer instead of a constraint might be better.
Regularization is widely used in the off-policy training.
\cite{fakoor2020p3o,brandfonbrener2021offline} proposed off-policy variants of TRPO, 
emphasizing the importance of regularization in bias-fixing.
GAIL~\cite{ho2016generative} emphasize the use of TRPO in their highly noisy training environment.
Our discussions and analysis of the bias-fixing  of regularized policy gradient methods are motivated by these studies.  

\section{Preliminaries}
\subsection{Markov Decision Process}
We consider the policy optimization for an episodic Markov decision process (MDP), $\mathcal{M} = \langle \mathcal{S}, \mathcal{A}, P, R, \gamma \rangle$,
  where $\mathcal{S}$ is the state space, $\mathcal{A}$ is the action space, $P:\mathcal{S} \times \mathcal{A} \rightarrow \Delta\mathcal{S}$ is the transition probability,
  $R:\mathcal{S} \times \mathcal{A} \rightarrow \mathbb{R}$ is the reward function, and $\gamma\in[0,1)$ is the discount factor.
Each episode ends when the agent enters the terminal absorbing state $T$, where the agent cannot leave and always receives zero rewards thereafter.
In the episodic setting, $\lim_{k\rightarrow \infty}\Pr(s_k=T)=1$.
A policy is a mapping $\pi_\theta(a|s):\mathcal{S} \rightarrow \Delta\mathcal{A}$, where $\theta$ is the parameter of this policy.
In DRL, $\theta$ is the weight of a neural network.

We have 
\begin{equation}
  d^1_\pi(s)= \sum_{k=0}^\infty\Pr(s_k=s|p_0, \pi)
\end{equation}
to denote the undiscounted state distribution induced by the policy $\pi$, where $p_0(s)$ is the initial state distribution.
We denote discounted state distribution using: 
\begin{equation}
  d^\gamma_\pi(s)= \sum_{k=0}^\infty\gamma^k \Pr(s_k=s|p_0, \pi).
\end{equation}
The undiscounted state distribution $d^1_{\pi}(s)$ is the state distribution when agents interact with the environment, 
 while the discounted $d^\gamma_{\pi}(s)$ is introduced in the derivation of policy gradients~\cite{sutton1999policy}.
For simplicity, we will use $d^\gamma_{\pi}$ and $d^1_{\pi}$ to denote these two state distributions $d^\gamma_{\pi}(s)$ and $d^1_{\pi}(s)$.

\subsection{Policy Optimization}

The policy optimization is to find an optimal policy $\pi^*_\theta(a|s)$ maximizing the expected discounted return $\rho(\pi_\theta)$, given by
\begin{equation}\label{eq:discounted}
  \rho(\pi_\theta):=\mathbb{E}_{\pi_\theta(a|s)}\left[\sum_{k=0}^{\infty} \gamma^k R\left(s_k, a_k\right)\right]
\end{equation}

\cite{sutton1999policy} have proved that the policy gradient of \cref{eq:discounted} can be written as
\begin{align}\label{eq:sutton}
  \nabla_\theta \rho(\pi)&=\mathbb{E}_{s\sim d^\gamma_{\pi}}\left[\sum_a \nabla_\theta\pi \cdot q_{\pi}\right],
\end{align}
where $q_{\pi}(s,a)$ is the action value function of policy $\pi(a|s)$,
  and we abuse the notation by using $\pi$ and $q_\pi$ to denote $\pi_\theta(a|s)$ and $q_\pi(s, a)$ respectively.

Eq. \ref{eq:sutton} can be cast as an iterative optimization style,
  where at each epoch we solve the following unified iterative optimization problem:
\begin{equation}\label{eq:1.optim}
  \pi_{t+1} = \argmax_{\pi} J_\gamma(\pi|\pi_t),
\end{equation}
with the objective function
\begin{align}\label{eq:obj_1}
  J_\gamma(\pi|\pi_t)&=\mathbb{E}_{s\sim d^\gamma_{\pi_t}}\mathbb E_{a\sim \pi} \left[q_{\pi_t}\right].
\end{align}
Starting from an initial policy $\pi_0$,
the iterative optimization procedure induces a sequence of policies $(\pi_0, \pi_1, \cdots)$.

However, common implementations of PG adopt the undiscounted state distribution $d^1_\pi$ instead of the discounted $d^\gamma_\pi$, leading to a biased objective function: 
\begin{equation}\label{eq:obj_undiscount}
  J_{1}(\pi|\pi_t)=\mathbb{E}_{s\sim d^1_{\pi_t}}\mathbb E_{a\sim \pi} \left[q_{\pi_t}\right].
\end{equation}

Apart from the objective function, we are also concerned with the optimization procedure.
SGD~\cite{zinkevich2003online} is one of the popular optimizers. 
At each step of the SGD optimization, a gradient $g$ of the objective function is attained based on one sample, and the parameter is updated with
\begin{equation}
  \theta \leftarrow \theta + \alpha g.
\end{equation}
Alternative choice of optimizer includes RMSProp~\cite{Tieleman2012RMSProp}.
It adapts the per-parameter learning rate by scaling it inversely proportional to the estimate of its second moment: 
\begin{equation}\label{eq:RMSProp}
  \theta_i \leftarrow \theta_i + \frac{\alpha}{\sqrt{G_{i}+\delta}} g_i,
\end{equation}
where $\theta_i$ is the $i$-th parameter, $g_i$ is the $i$-th gradient, $G_i$ estimates the squared gradient of the $\theta_i$, and $\delta$ is a small constant.

\subsection{Regularization}
Regularization such as KL divergence has been long used and studied in policy optimization~\cite{schulman2015trust,abdolmaleki2018maximum,nachum2018trustpcl,hessel2021muesli}. 
The augmented form of a general objective function is
\begin{equation}\label{eq:obj_reg}
  J(\pi|\pi_t) = \mathbb{E}_{s\sim d_\pi }\left\{\mathbb E_{a\sim \pi} \left[q_{\pi_t}\right]-\alpha \KL\left(\pi_t||\pi\right)\right\},
\end{equation}
where $\alpha$ is a hyperparameter that controls the strength of regularization.

To isolate the effect of regularization, we also derive the following form reflecting the influence of regularization:
\begin{equation}\label{eq:obj_entropy_derivation}
  \begin{split}
  &\argmax_\pi J(\pi|\pi_t) \\
  &= \argmax_\pi \mathbb{E}_{s\sim d_{\pi_t}}\left\{\mathbb E_{a\sim \pi} \left[q_{\pi_t}\right]-
      \alpha KL\left(\pi_t||\pi\right)\right\} \\
  &= \argmax_\pi \mathbb{E}_{s\sim d_{\pi_t}}\{\mathbb E_{a\sim \pi} \left[q_{\pi_t}\right] + \alpha\mathbb E_{a\sim \pi_t}\left[\log \frac{\pi}{\pi_t}\right]\}\\
  &= \argmax_\pi \mathbb{E}_{s\sim d_{\pi_t}}\mathbb E_{a\sim \pi_t} 
      \left[ \frac{\pi}{\pi_t}{q}_{\pi_t}+\alpha\log \pi\right].
  \end{split}
\end{equation}
We drop a constant term in the last equation.
This objective function can be easily implemented by adding an entropy term to the original Q value.
The derivation of the reverse KL is similar, which is provided in \cref{appendix: reverse_kl}.

\section{Bias in Policy Gradient Estimation}\label{sec:bias}
In this section, we discuss a scenario where using the biased policy gradient \cref{eq:obj_undiscount} leads to a highly sub-optimal outcome compared with the unbiased version \cref{eq:obj_1}.

We first introduce an example called \textit{state alias}.

\begin{figure}[t]
  \centering
  \includegraphics[width=0.9\linewidth]{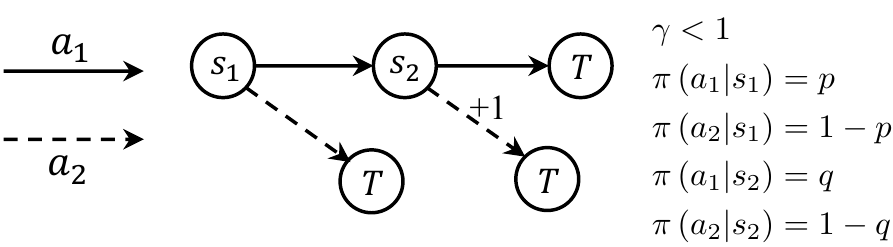}
  \caption{
    A \textit{state alias} example where the undiscounted objective function leads to a much worse result.
  }\label{fig:alias}
\end{figure}

In~\Cref{fig:alias}, the agent is initialized at $s_1$ in each episode and only receives a $+1$ reward when it chooses action $a_2$ in state $s_2$. The episode ends when it reaches state $T$.
State alias phenomenon emerges when it is hard for the agent to tell the differences between the two states, thus these two states have to share similar policy distributions:
\begin{equation}\label{eq:alias_constraint}
  \max_a \| \pi(a|s_1) - \pi(a|s_2) \| \leq \epsilon, \epsilon < \gamma.
\end{equation}
Suppose $\pi(a_1|s_1)=p$ and $\pi(a_1|s_2)=q$. The expected discounted return of this example is $\rho(p, q) = p (1-q) \gamma$.
To maximize the return, we want $p$ to be large and $q$ to be small while respecting the constraint \cref{eq:alias_constraint}, so we let $q=p-\epsilon$.
In this example, the only globally optimal policy is $p^*=\frac{1+\epsilon}{2}, q^* =\frac{1-\epsilon}{2}$, regardless of the choice of $\gamma$ and with the maximum expected reward 
\begin{equation}
  \rho(p^*, q^*) = \frac{(1+\epsilon)^2}{4}\gamma.
\end{equation}
Note this is the \textit{only} rational policy, thus any algorithms and implementations leading to other policies are questionable.
However, when we simulate the procedure of policy optimization, sub-optimal result occurs.

Given some $p$ and $q=p-\epsilon$, we compute the gradient of each state based on the advantages of choosing $a_1$ in two states:
\begin{equation}
  g(s_1|p) = (1-q)\gamma = (1-p+\epsilon)\gamma \qquad g(s_2|p) = -1.
\end{equation}
The discounted (unbiased) state distribution is
\begin{equation*}
  d_\gamma(s_1|p) = 1 \qquad d_\gamma(s_2|p) = \gamma p,
\end{equation*}
while the undiscounted (biased) state distribution is
\begin{equation*}
  d_1(s_1|p) = 1 \qquad d_1(s_2|p) = p.
\end{equation*}
The total gradient on the discounted and undiscounted state distribution settings is given by
\begin{align}
  &g_\gamma(p) = \sum_{s\in S} g(s|p) \cdot d_\gamma(s|p) = (1-p+\epsilon) \gamma- \gamma p, \\
  &g_1(p) = (1-p+\epsilon) \gamma- p.
\end{align}
When the gradients $g_\gamma$ and $g_1$ vanish, we attain the converged policies for two state distributions and their corresponding expected returns:
\begin{align}
  p_\gamma^* &= \frac{1+\epsilon}{2} 
  &q_\gamma^* &= \frac{1-\epsilon}{2} 
  &\rho(p_\gamma^*, q_\gamma^*) &= \frac{(1+\epsilon)^2}{4}\gamma
  \\
  p_1^* &= \frac{\gamma(1+\epsilon)}{1+\gamma} 
  &q_1^* &= \frac{\gamma-\epsilon}{1+\gamma}  
  &\rho(p_1^*, q_1^*) &= \frac{(1+\epsilon)^2\gamma^2}{(1+\gamma)^2}.
\end{align}
The unbiased PG is consistent with the theoretically optimal solution, while the biased PG leads to a much worse result with a performance decay
\begin{equation}
  \frac{\rho(p_1^*, q_1^*)}{\rho(p^*, q^*)} = \frac{4\gamma}{(1+\gamma)^2} < 1.
\end{equation}
Thus, the biased PG always leads to a suboptimal outcome when the state alias phenomenon occurs.

When $\epsilon=0$, the numerical example degenerates to the one described in \cite{nota2020policy}. 
We describe it in detail in \cref{appendix: alias}.

State alias phenomenon will occur in DRL under the neural network parameterization when it maps the high-dimension states into the low-dimension feature representations.
Some similar state alias phenomenons have been widely observed and discussed in the literature.
Representation learning methods like~\cite{lehnert2020successor,uehara2022representation,li2021fault} compressed high-dimensional observations input into low-dimensional state space, naturally leading to the risk of state alias theoretically. 
\cite{agarwal2021contrastive} conducted a contrastive learning method to distinguish similar states to attain a better policy.
\cite{frazier2019improving} discussed the severe state alias problem in complicated 3D virtual environments like the game Minecraft.
\cite{liu2021returnbased} achieved a better outcome by distinguishing similar state-action pairs.

\begin{figure}[t]
  \centering
  \begin{minipage}[t]{0.42\linewidth}
    \centering
    \includegraphics[width=\linewidth]{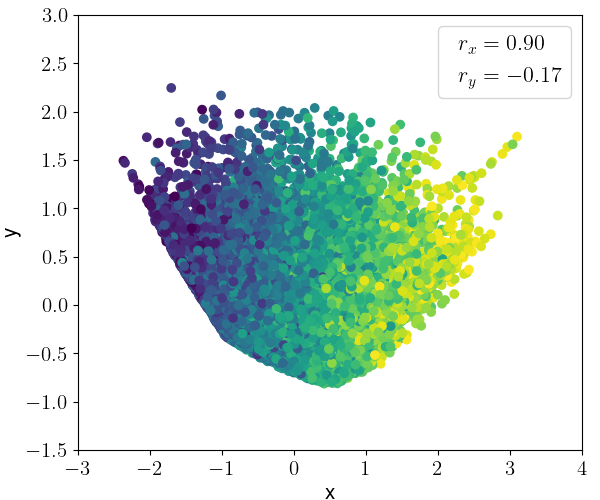}
    \subcaption{Optimal}\label{fig:alias_nn_a}
  \end{minipage}
  \begin{minipage}[t]{0.42\linewidth}
      \centering
      \includegraphics[width=\linewidth]{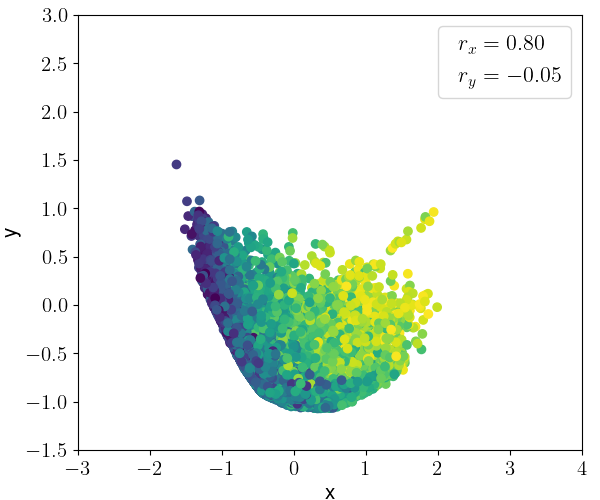}
      \subcaption{Early stage}\label{fig:alias_nn_b}
  \end{minipage}
  \caption{
    Feature representations at different stages of training.
    With the same dataset of $\{(s, a)\}$ pairs, we compute the feature representation $f(s)$ by passing the state through the first two layers of the neural network.
    We visualize the $\{f(s)\}$ on a 2D plane with PCA.
    Color represents the value of the real action (lighter color means higher value).
    In the early stage of training, the feature representation erroneously clusters the states with different action values.
    The absolute correlation coefficients of two dimensions ($r_x$ and $r_y$) are provided in the figures.
  }\label{fig:alias_nn}
\end{figure}

A similar state alias phenomenon is observed in our experiment in the Inverted Pendulum environment~\cite{brockman2016openai}, especially in the early stages.
We visualize the feature representation (passing observation over the first two feature layers of the NN) by mapping it into a 2D surface with Principal Component Analysis (PCA) in \cref{fig:alias_nn}.
The color represents the true action value (as the action space is 1D continuous); we hope feature layers cluster states with similar action values, otherwise the state alias is more likely to occur: the feature layer erroneously maps the states which should have different policy distributions to the similar feature representations.
We visualize the feature representations of the same dataset at different stages of training.
\cref{fig:alias_nn_a} is the visualization of state feature representations of the optimal policy, acting as the ground truth, while \cref{fig:alias_nn_b} is in the early stage of training.
In the early stage of training, the states are erroneously gathered and mixed up with different action values, indicating the high risk of state alias.
The absolute Pearson correlation coefficients between actions and feature representations are smaller in the early stages, consistent with the visualization.
A detailed description and more results for this experiment can be found in \cref{appendix:c}.

\section{Bias Rectification}\label{sec:rectification}
We discussed a phenomenon where the undiscounted bias leads to a highly sub-optimal result, and pointed out that it is likely to occur during the training of DRL.
Then why does the empirically biased version of PG methods still work well in practice?
We discover three common but overlooked techniques that notably help mitigate the influence of bias: 1.\ a proper learning rate; 2.\ adaptive learning rate optimizers; 3.\ regularization.
Experiments to support our findings are provided in \cref{fig:exp}, with detailed explainations in \cref{sec:exp}.

\subsection{Learning Rate}
A common perception across the area of DRL is that a proper learning rate is beneficial to the stability of the model training process~\cite{goodfellow2016deep}.
We further find a relatively small learning rate is also beneficial for bias rectification.

However, a smaller learning rate directly leads to a much slower convergence process, 
  and tuning the learning rate to balance the stability and convergence speed is a tedious task.
Large-scale empirical studies~\cite{andrychowicz2021what,engstrom2019implementation} show that the optimal learning rate varies across different environments and algorithms.
Thus, a high learning rate tuning reliable method is less welcomed.

\subsection{Optimizer}
Motivated by the importance of learning rate, we investigate the effect of adaptive-learning-rate-based optimizers.
We find the widely used adaptive-learning-rate-based optimizers, such as RMSProp~\cite{Tieleman2012RMSProp} and Adam~\cite{kingma2014adam}, help mitigate the harm of biases as well.
As Adam can be viewed as a combination of RMSProp and momentum~\cite{kingma2014adam}, we use RMSProp as an example, with an update rule in \cref{eq:RMSProp}.

RMSProp estimates the second moment of the gradients by statistics, and divides the parameter gradient by the square root of the second moment.
The effect of RMSProp can be explained in two ways.

First, it mitigates the effects of the model oscillations that are caused by biased PG estimates.
RMSProp restricts the vertical changes, thus reducing the possibility of collapsing.

Second, when optimizing the policy distribution, RMSProp can be seen as estimating an empirical Fisher information matrix (FIM), which is a measure of the curvature of the loss surface.
The element in the FIM of the objective function $F(\theta)$ is defined as follows~\cite{amari2012differential}:
\begin{equation}\label{eq:Fisher}
  F_{i, j}(\theta) = \mathbb E_{s\sim d^\gamma_\pi} \mathbb E_{a\sim \pi}\left[\frac{\partial \log \pi}{\partial \theta_i} \frac{\partial \log \pi}{\partial \theta_j}\right].
\end{equation}
We can rewrite the update rule of RMSProp in \cref{eq:RMSProp} to a matrix form:
\begin{equation}\label{eq:RMSProp_matrix}
  \theta \leftarrow \theta + \bar{F}^{-1}(\theta)\nabla_\theta J(\theta),
\end{equation}
where $\bar{F}(\theta)$ is the diagonal matrix with its $i$-th diagonal element being an estimate of the second moment of the gradient:
\begin{equation}\label{eq:empirical_fisher}
  \bar{F}_{i, i}(\theta) = \mathbb E_{s\sim d^\gamma_\pi} \mathbb E_{a \sim \pi}\left[\textup{Diag}\left[\frac{\partial \log \pi}{\partial \theta_i} \frac{\partial \log \pi}{\partial \theta_i}\right]\right].
\end{equation}
It is the diagonal approximation of the FIM.
Thanks to the sparsity of the gradient under NN parameterization, the inverse of the diagonal approximation does not lose much information~\cite{zeiler2012adadelta,becker1988improving}.
The empirical FIM is a key factor to stabilize the training process.

As for the other popular optimizer, Adam~\cite{kingma2014adam}, is a combination of RMSProp and Momentum. Adam has a similar bias rectification ability.
However, using Momentum~\cite{goodfellow2016deep} only without the help of RMSProp does not help much.
We omit the discussion, but give the result in \cref{fig:exp_b}.

\subsection{Regularization}\label{subsec:regu}
\begin{figure}[t]
  \centering
  \includegraphics[width=0.75\linewidth]{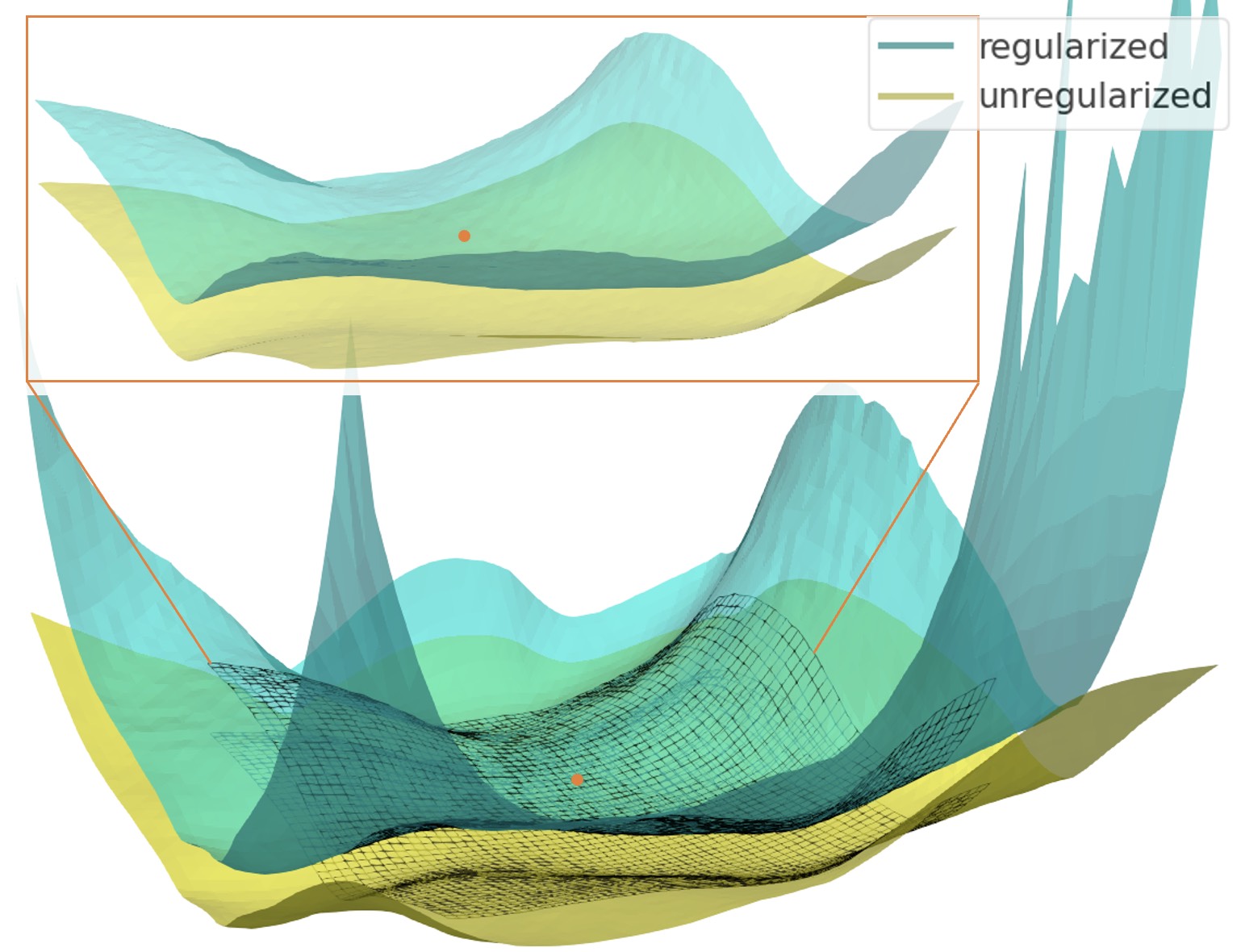}
  \caption{
    The blue and yellow surfaces are the loss surfaces with and without regularization at epoch 20, respectively. 
    The center dark area is zoomed in in the orange box.
    The orange dot indicates the position of the current policy.  
  }\label{fig:loss_surface}
\end{figure}

The derivation in \cref{eq:Fisher,eq:RMSProp_matrix,eq:empirical_fisher} reveals the close connection between RMSProp and FIM-based methods.
As the approximate analytic solution of KL regularization constrained method like TRPO~\cite{schulman2015trust} follows the form of \cref{eq:RMSProp_matrix}~\cite{agarwal2021reinforcement}, we investigate the effect of KL regularization.

We discover that both the KL and the reverse KL regularization are useful for bias rectification.
We adopt the simple formulation in \cref{eq:obj_entropy_derivation} for KL regularization,
  and conduct our experiments with the following objective function:
\begin{equation}\label{eq:kl}
  J(\pi|\pi_t) = \mathbb{E}_{s\sim d_{\pi_t}}\mathbb E_{a\sim \pi_t} 
  \left[ \frac{\pi}{\pi_t}{q}_{\pi_t}+\alpha\log \pi\right].
\end{equation}
The objective function for reverse KL regularization:
\begin{align}\label{eq:reverse_kl}
  J(\pi|\pi_t) = \mathbb{E}_{s\sim d_{\pi_t}}&\mathbb E_{a\sim \pi_t} 
  \left[\frac{\pi}{\pi_t}q_{\pi_t}+\beta\frac{\pi}{\pi_t}\left(\log\pi_t-\log\pi\right)\right].
\end{align}
The derivation is similar to the KL regularization. We put it in \cref{appendix: reverse_kl}.

To explain the effect of regularization, we plot the loss surfaces of the objective function with and without regularization at epoch 20 in \cref{fig:loss_surface} with the visualization method in~\cite{li2018visualizing}.
In the small neighborhood of the current policy, the loss surface with regularization is similar to the one without regularization.
However, the loss surface rapidly becomes harsh in the larger neighborhood in the regularized case.
This ``potential well'' mechanism maintains the performance, mitigating large disturbances caused by biased gradient estimations.
The experiment details are provided in \cref{appendix:d2}.

\section{Experiments} \label{sec:exp}
\subsection{Experimental Setup} \label{sec:exp_design}
Our experiments are conducted on the Mujoco environment in the OpenAI Gym~\cite{brockman2016openai,todorov2012mujoco}.
We simplify the experiments by truncating the episodes at 200 steps.

To exclude the effect of some modern techniques and code-level tricks, 
we use a unified approach, with a simple network,
  and estimate the value function with a Monte Carlo method~\cite{sutton2018reinforcement}.
Monte Carlo methods give an unbiased estimate of the value function, thus we can focus on the influence of the bias of the actor.
We use the SGD~\cite{zinkevich2003online} as the baseline optimizer, with a learning rate that decays every several epochs.
In each epoch, we sample a batch of trajectories from the current policy and update the policy.

We conduct the following two experiments:

1) \textit{Performance Experiment}. 
This is an experiment to compare the performances of algorithms. 
We test how our experimental methods such as low learning rate, RMSProp and KL regularization influence the performance of biased PG.
We train an unbiased baseline with the objective function \cref{eq:obj_1} and the iterative scheme in \cref{eq:1.optim}.
We train three other models with the same hyperparameter: biased+baseline, unbiased+experimental and biased+experimental, where experimental stands for any one of the aforementioned rectification techniques.

\begin{figure}[t]
  \centering
  \includegraphics[width=0.99\linewidth]{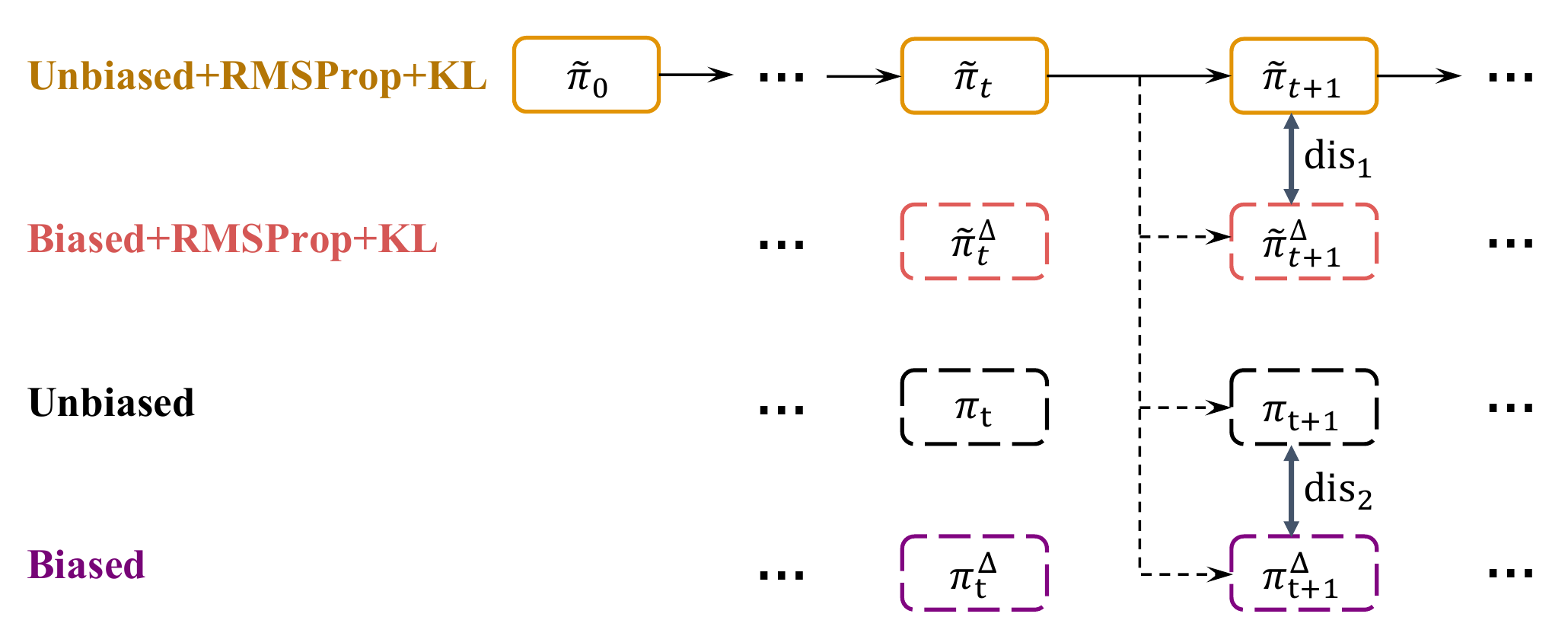}
  \caption{Bias Spread Experiment Design.
    At each epoch $t$, we train 4 different models based on the same start policy $\pi_t$.
    Then we compute the distance based on \cref{eq:EARD} between the unbiased and biased pairs.}\label{fig:bias_spread}
\end{figure}

2) \textit{Bias Spread Experiment.}
To analyze the influence of bias along the training procedure, we design our experiments as illusrated in~\Cref{fig:bias_spread}.
At each epoch $t$, with a starting model initialized with the baseline model $\pi_t$, we train 4 different models with the same hyperparameter and dataset, but with different methods: unbiased, biased, unbiased+RMSProp+KL, and biased+RMSProp+KL.

After training, we compute the model distance $\dis$ between biased and unbiased models to measure the influence of bias on the training.
The expected absolute ratio deviation (EARD)~\cite{sun2022you}, is adopted as a measure of model distance $\dis$.
The EARD is defined in the following.
\begin{definition}[expected absolute ratio deviation (EARD)]\label{def:EARD}
  Suppose $\pi_{t+1}$ and $\pi^\Delta_{t+1}$ are two policies trained from $\pi_t$ with parameters $\theta_1$ and $\theta_2$, respectively.
  The EARD measures the distance between these two models:
  \begin{equation}\label{eq:EARD}
    \dis = \mathbb{E}_{s \sim d_{\pi_t}}\mathbb{E}_{a \sim \pi_t}\left|\frac{\pi_{t+1}(a \mid s)}{\pi^\Delta_{t+1}(a \mid s)}-1\right|.
  \end{equation}
\end{definition}

The distance measures the influence of bias at epoch $t$. 
We use this strategy to probe the bias spread along the training process.
A smaller EARD distance indicates a smaller influence of bias.

\subsection{Undiscounted Bias Rectification}
\begin{figure*}[ht]
  \begin{minipage}[t]{\linewidth}
    \centering
    \begin{minipage}[t]{0.33\linewidth}
      \centering
      \includegraphics[width=0.9\linewidth]{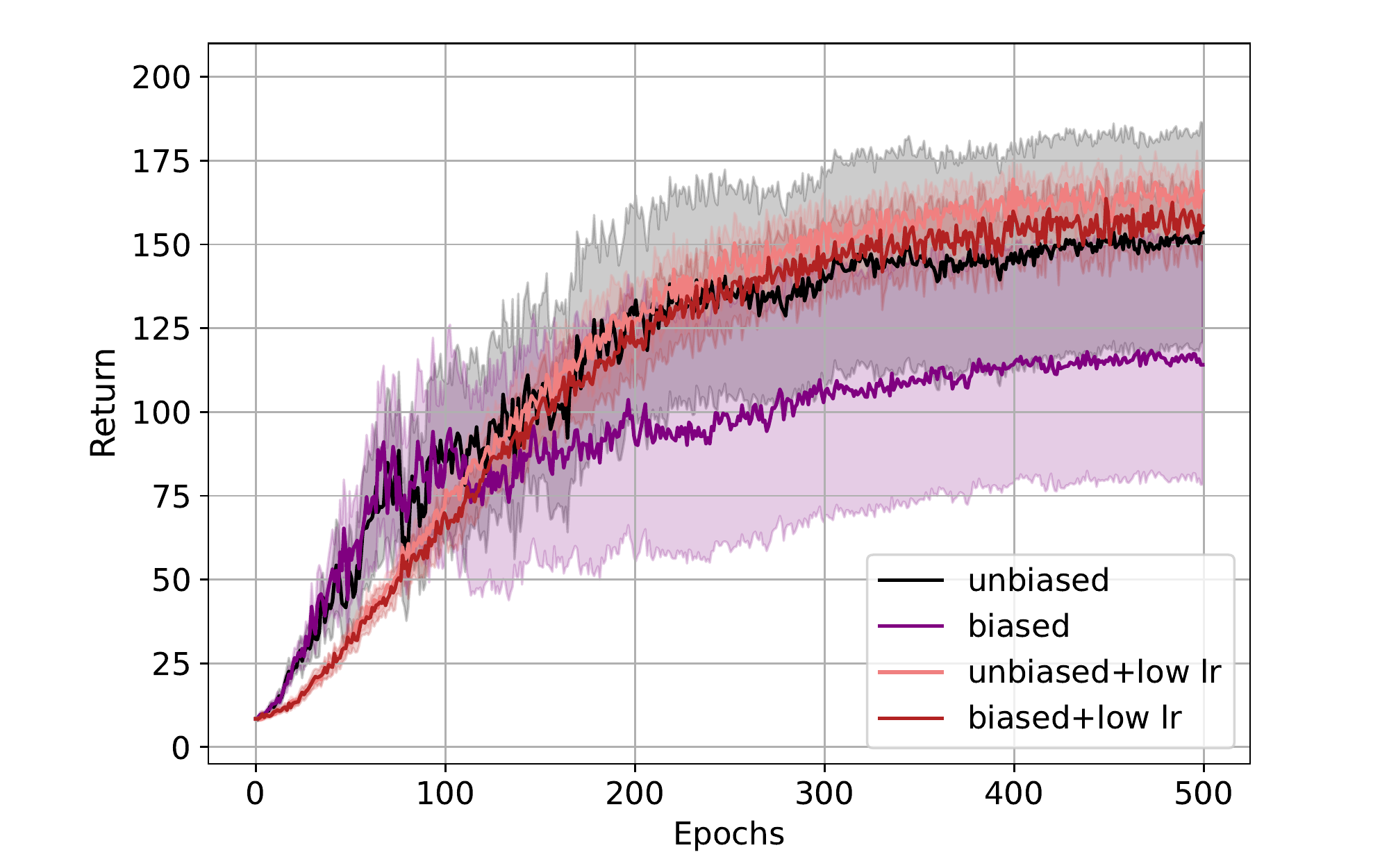}
      \subcaption{Low learning rate on undiscounted bias}\label{fig:exp_a}
    \end{minipage}
    \begin{minipage}[t]{0.33\linewidth}
        \centering
        \includegraphics[width=0.9\linewidth]{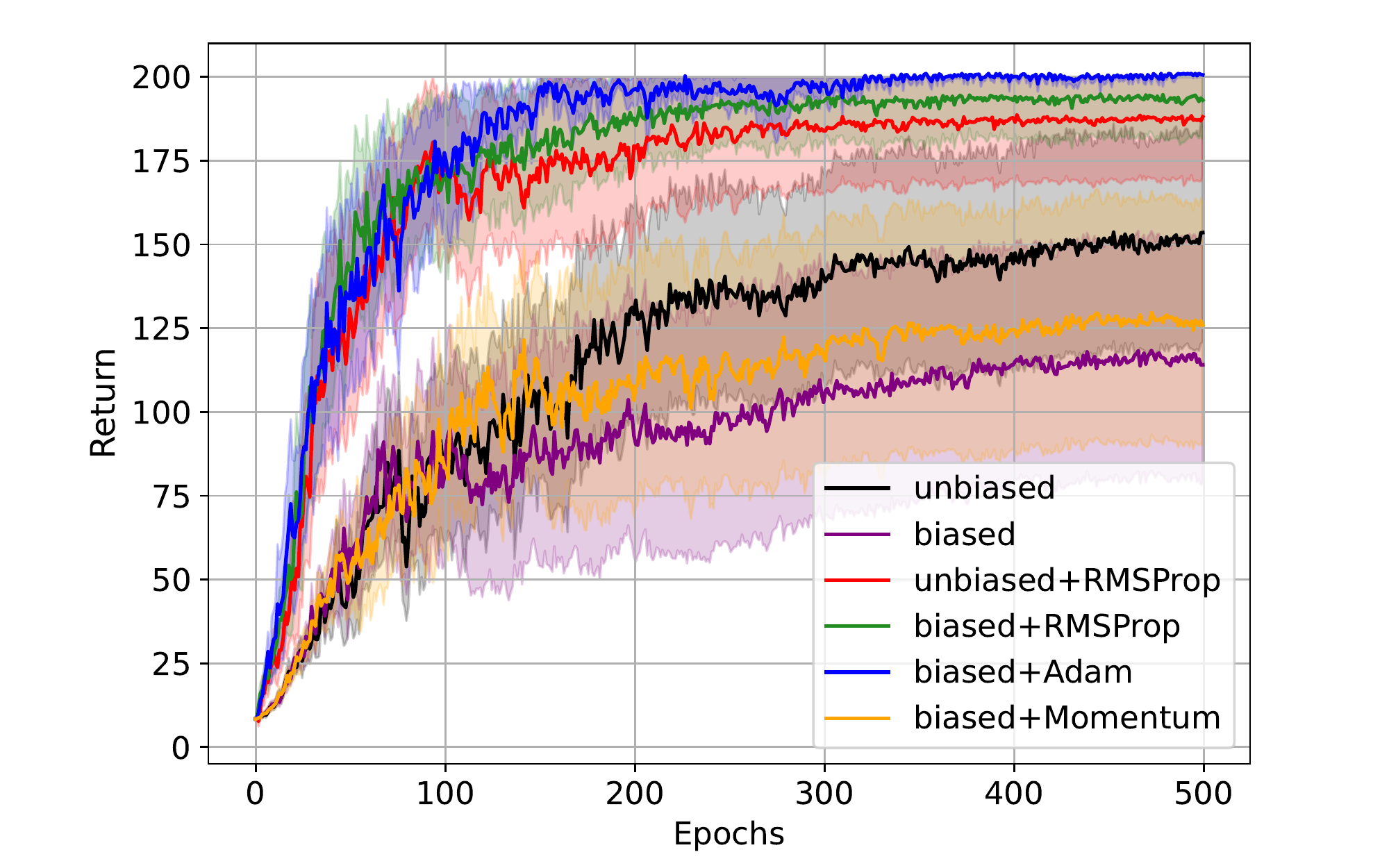}
        \subcaption{RMSProp on undiscounted bias}\label{fig:exp_b}
    \end{minipage}
    \begin{minipage}[t]{0.33\linewidth}
        \centering
        \includegraphics[width=0.9\linewidth]{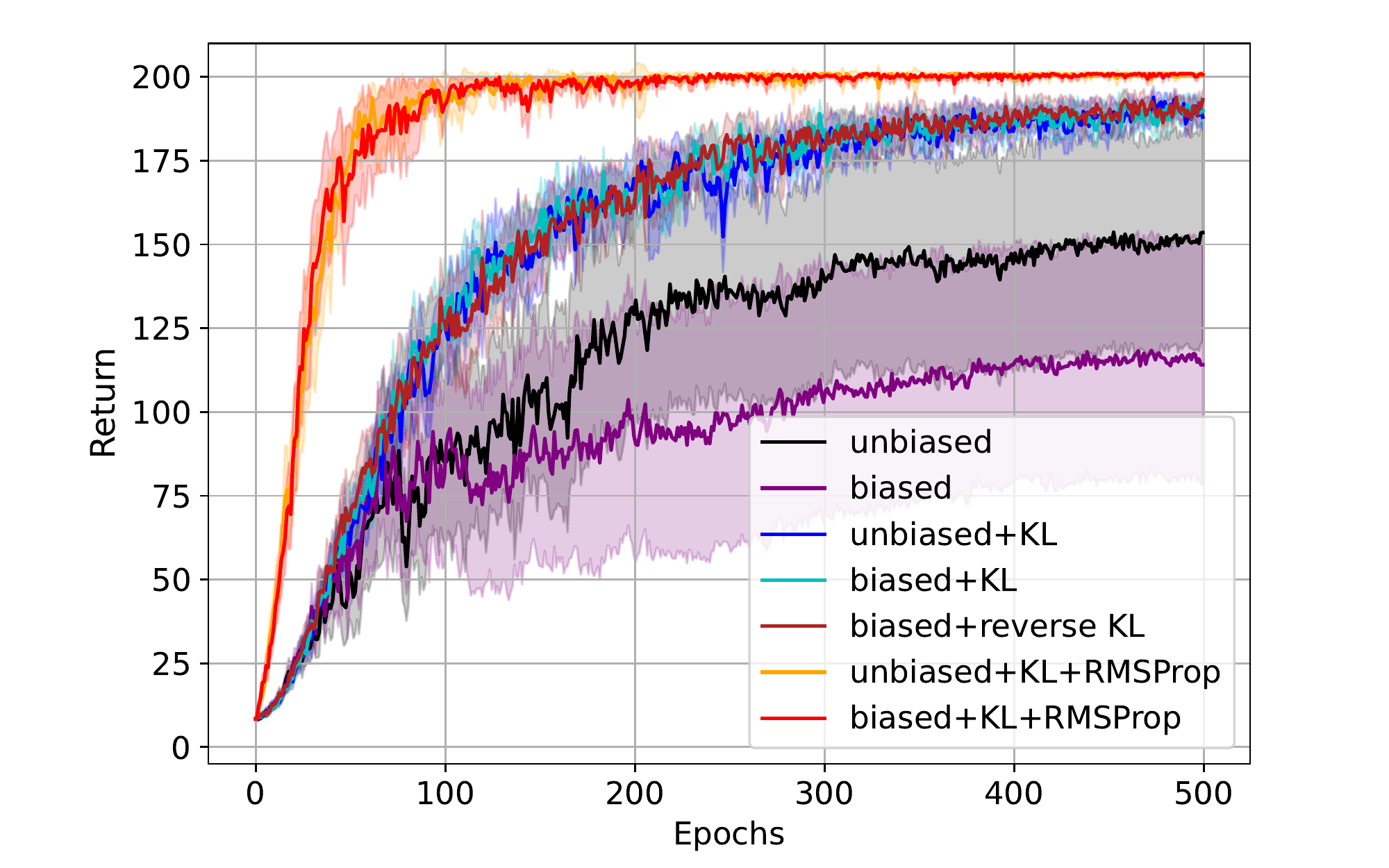}
        \subcaption{KL regularization on undiscounted bias}\label{fig:exp_c}
    \end{minipage}
  \end{minipage}
  \caption{
    Performance experiments comparing three experimental techniques with the undiscounted bias (20 random seeds). 
  }\label{fig:exp}
\end{figure*}

\begin{figure*}[t]
  \begin{minipage}[t]{\linewidth}
    \centering
    \begin{minipage}[t]{0.19\linewidth}
        \begin{minipage}[t]{\linewidth}
          \includegraphics[width=\linewidth]{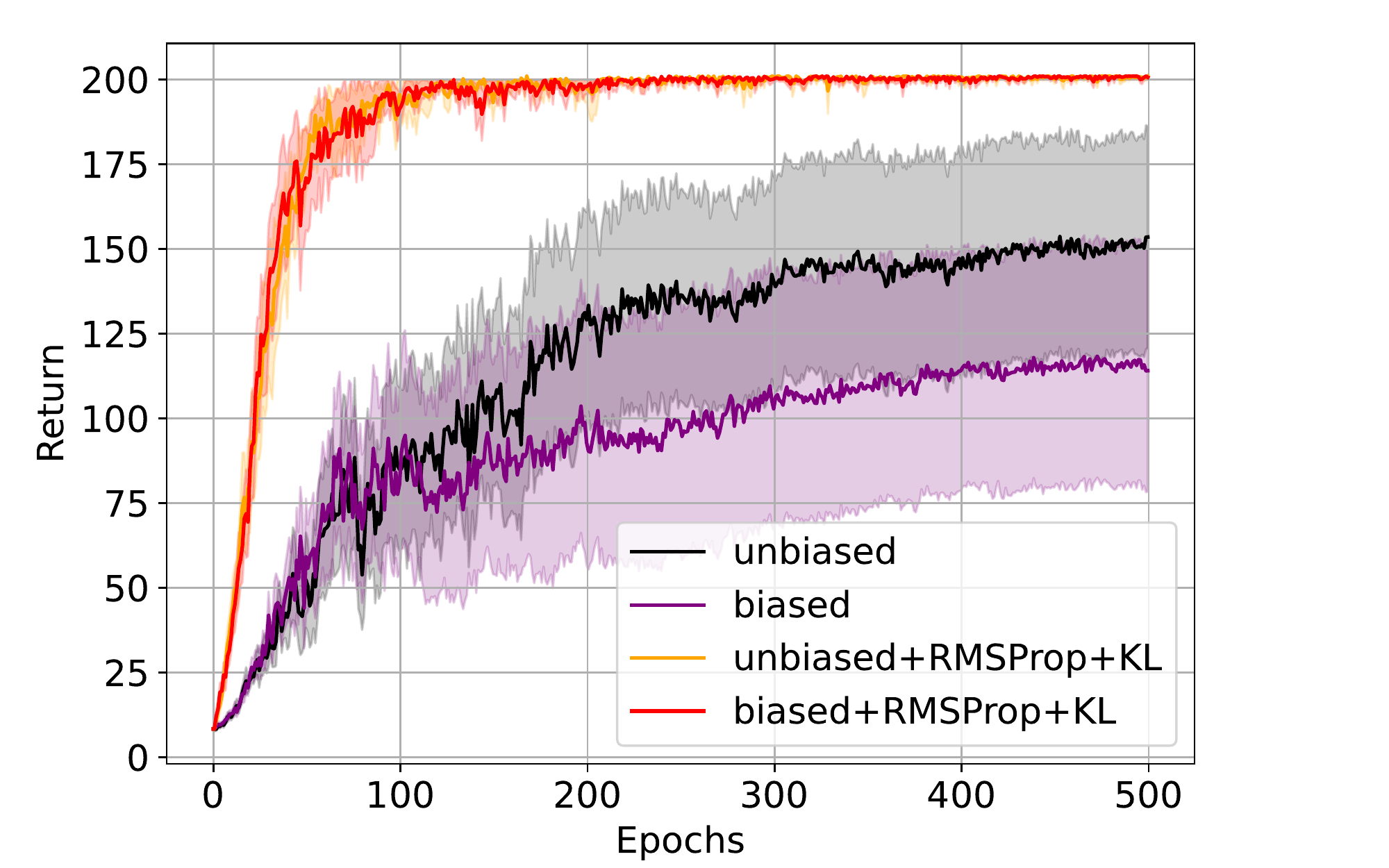}
        \end{minipage}
        \begin{minipage}[t]{\linewidth}
          \includegraphics[width=\linewidth]{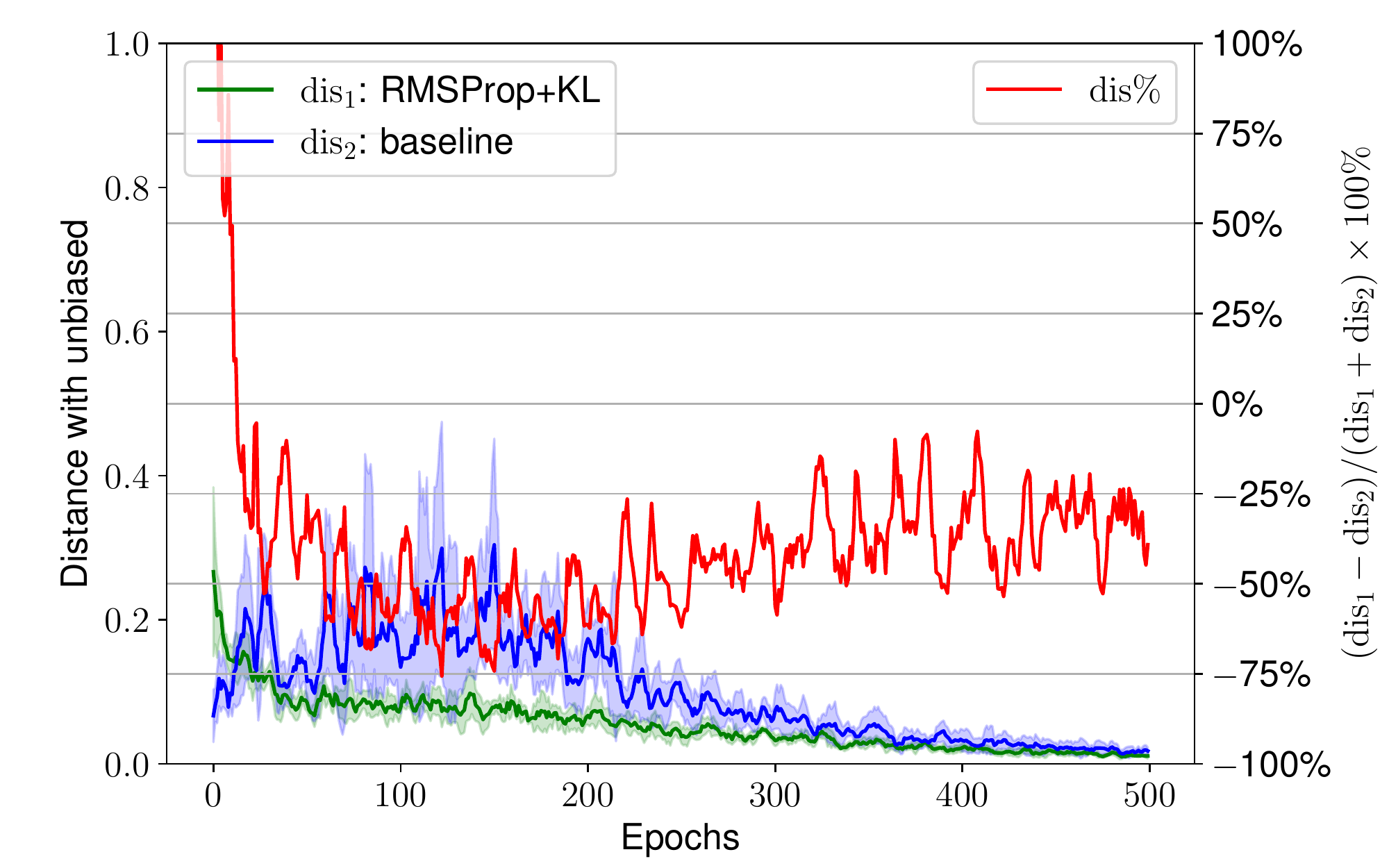}
        \end{minipage}
      \subcaption{Inverted Pendulum}
    \end{minipage}
    \begin{minipage}[t]{0.19\linewidth}
      \begin{minipage}[t]{\linewidth}
        \includegraphics[width=\linewidth]{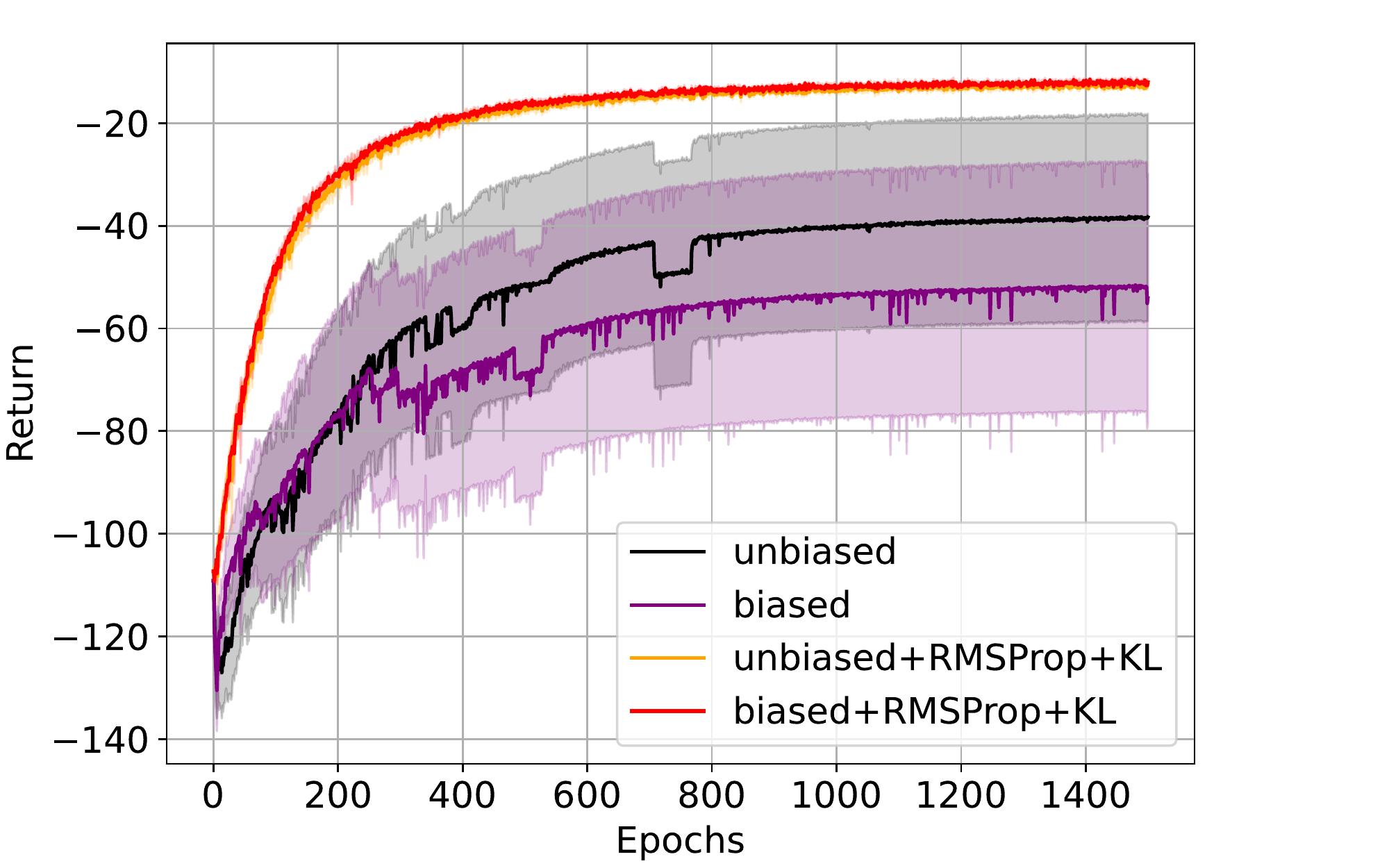}
      \end{minipage}
      \begin{minipage}[t]{\linewidth}
        \includegraphics[width=\linewidth]{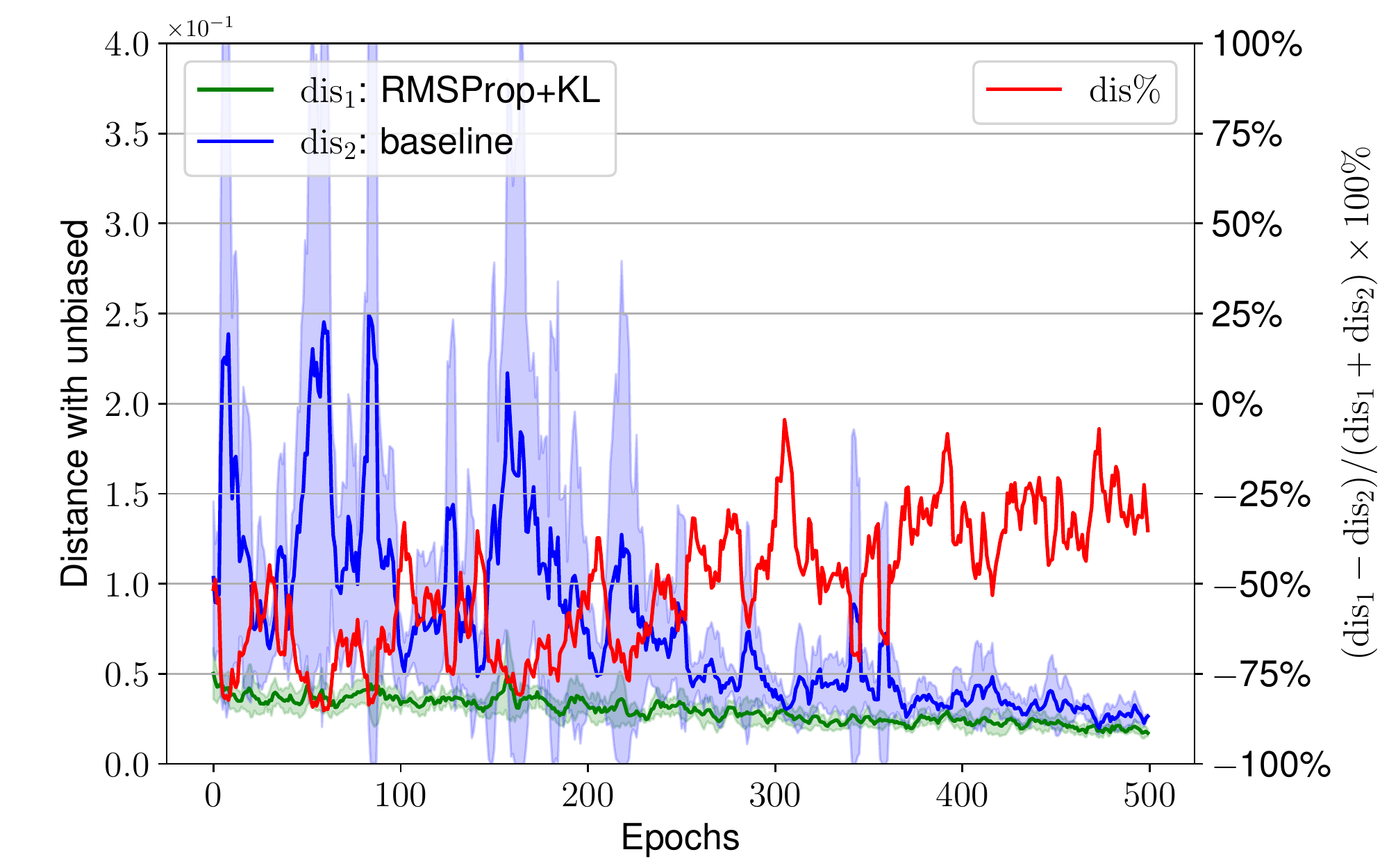}
      \end{minipage}
      \subcaption{Reacher}
    \end{minipage}
    \begin{minipage}[t]{0.19\linewidth}
      \begin{minipage}[t]{\linewidth}
        \includegraphics[width=\linewidth]{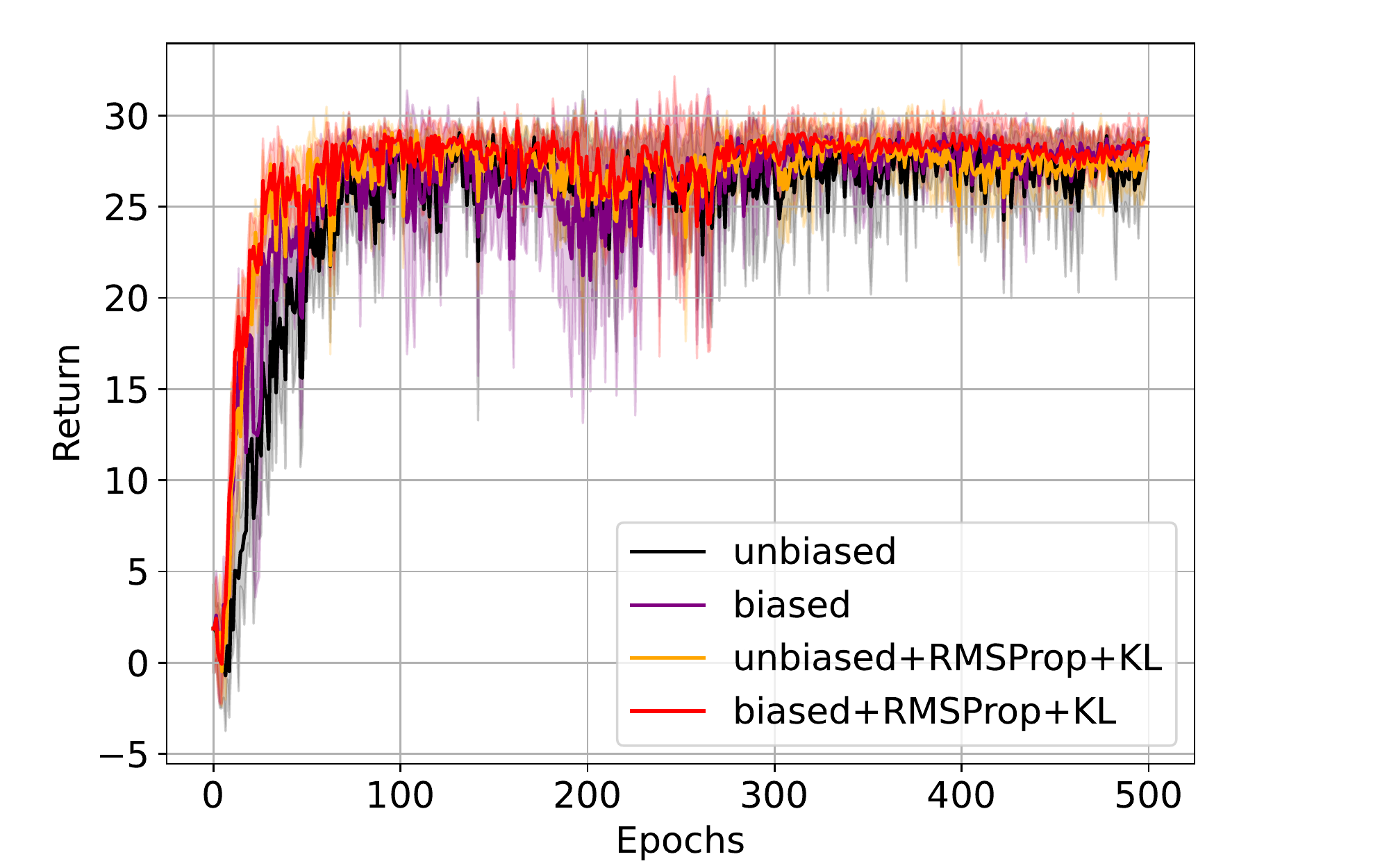}
      \end{minipage}
      \begin{minipage}[t]{\linewidth}
        \includegraphics[width=\linewidth]{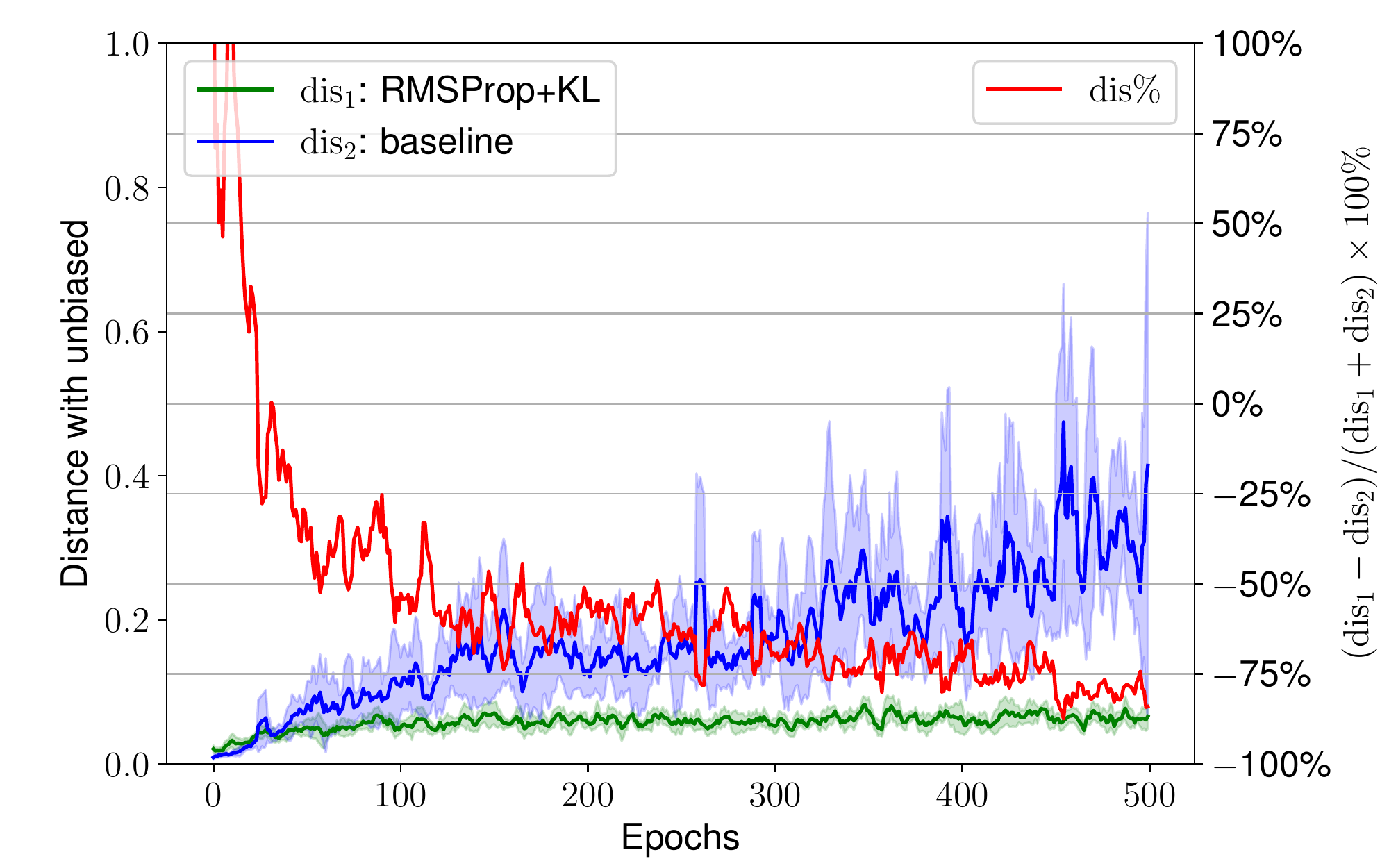}
      \end{minipage}
      \subcaption{Swimmer}
    \end{minipage}
    \begin{minipage}[t]{0.19\linewidth}
      \begin{minipage}[t]{\linewidth}
        \includegraphics[width=\linewidth]{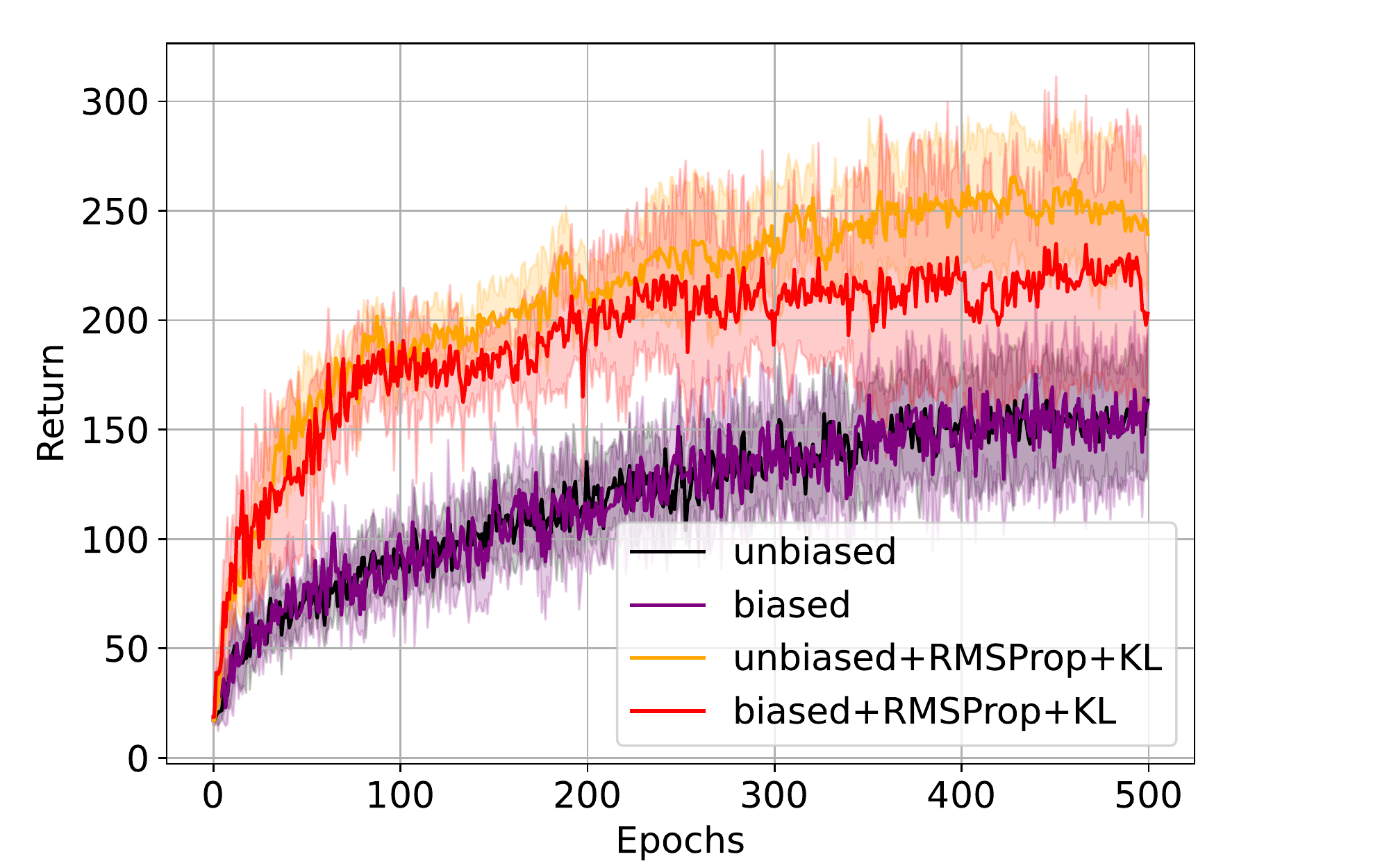}
      \end{minipage}
      \begin{minipage}[t]{\linewidth}
        \includegraphics[width=\linewidth]{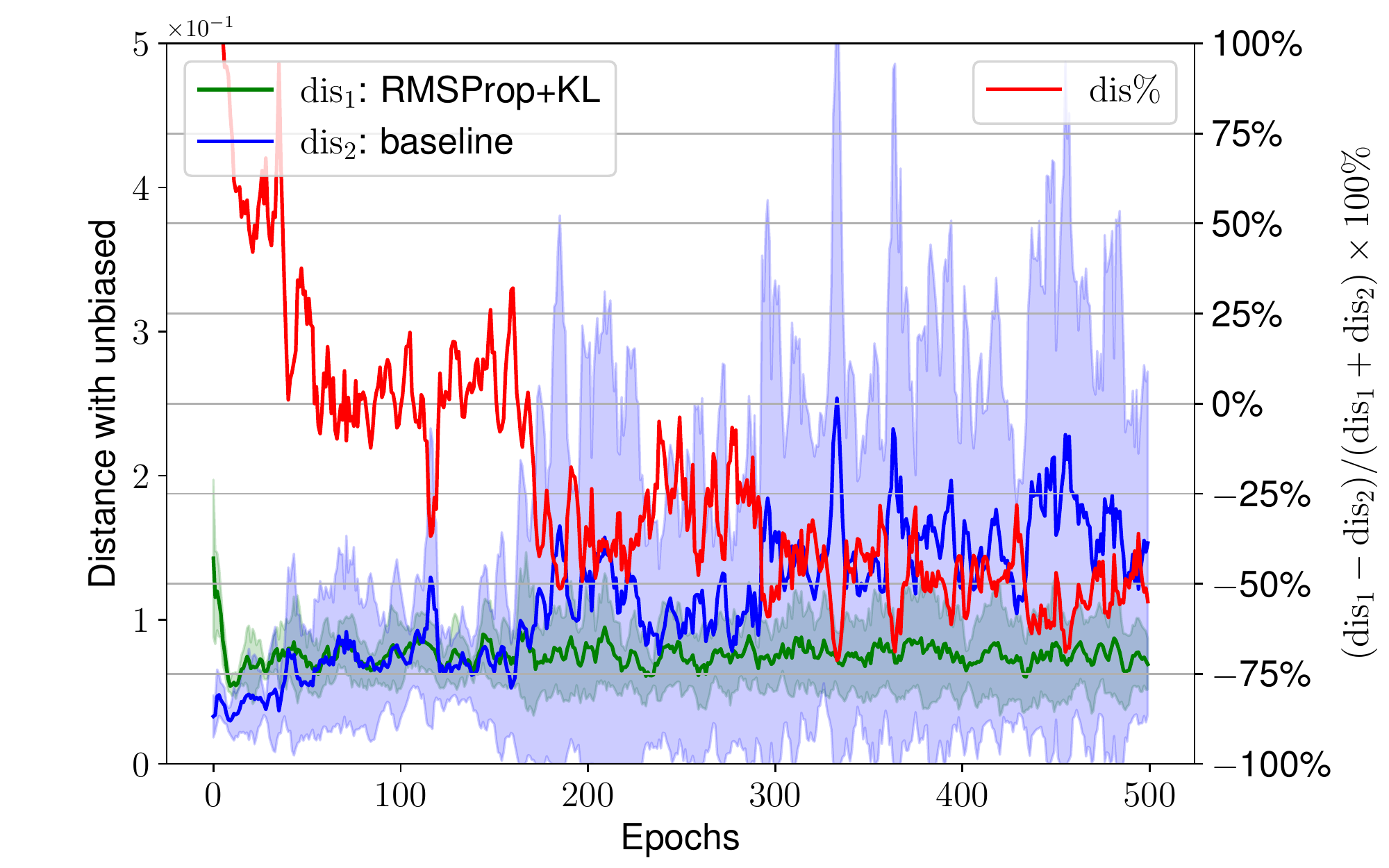}
      \end{minipage}
      \subcaption{Hopper}
    \end{minipage}
    \begin{minipage}[t]{0.19\linewidth}
      \captionsetup{justification=centering}
      \begin{minipage}[t]{\linewidth}
        \includegraphics[width=\linewidth]{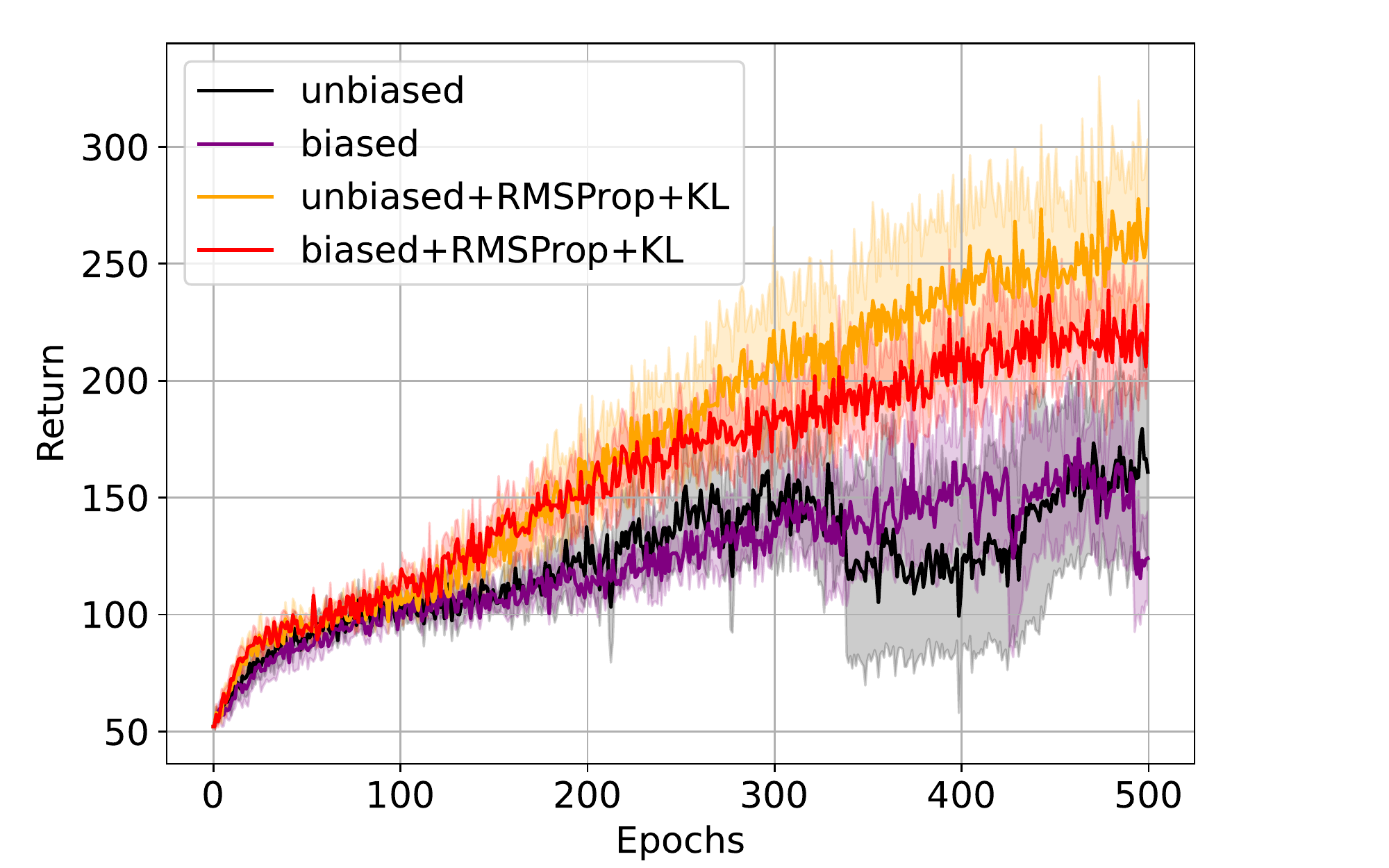}
      \end{minipage}
      \begin{minipage}[t]{\linewidth}
        \includegraphics[width=\linewidth]{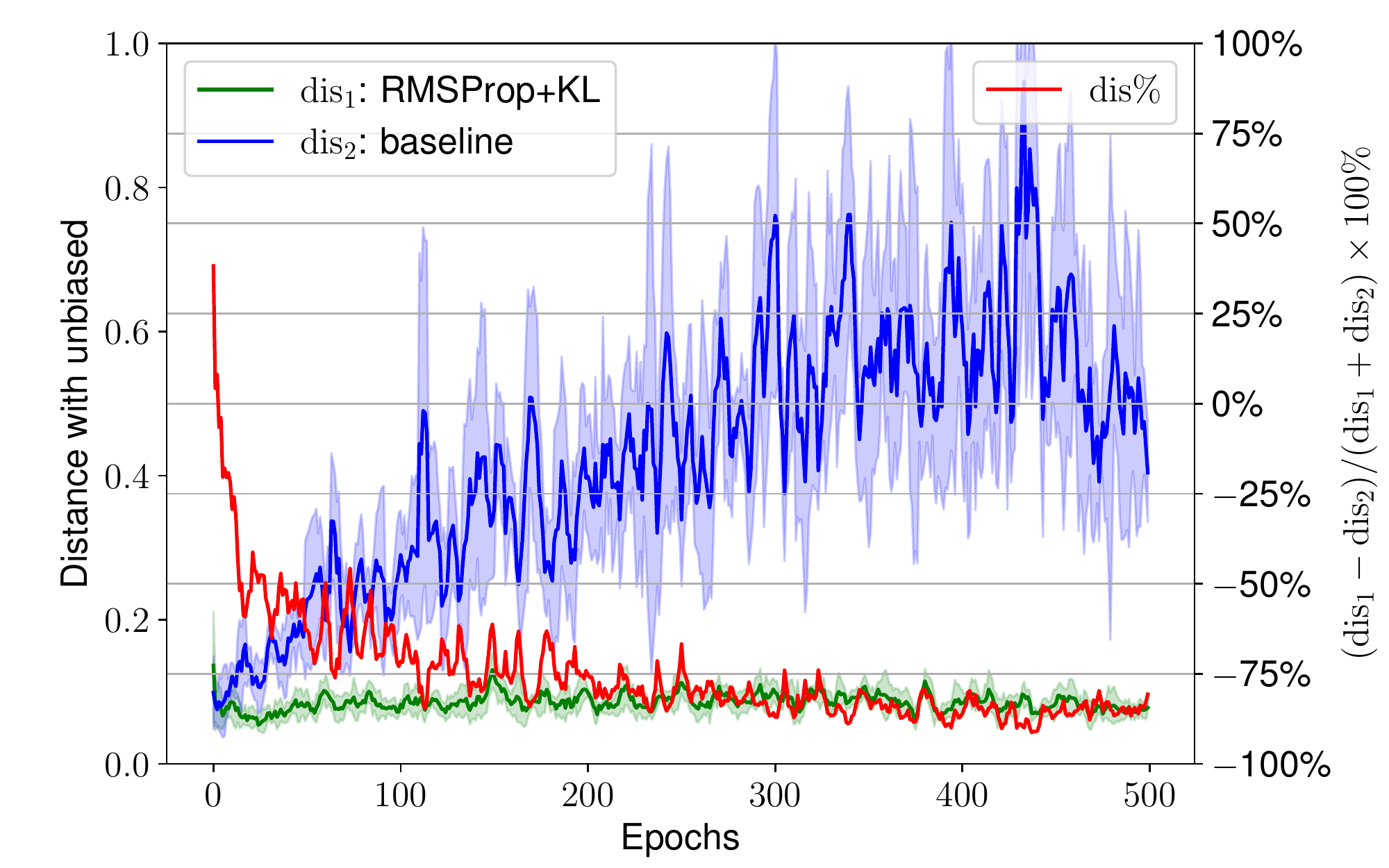}
      \end{minipage}
      \subcaption{Inverted\\Double Pendulum}
    \end{minipage}
  \end{minipage}
  \caption{
    Bias rectification in the continuous control Mujoco environment (5 random seeds). 
    The top row is the performance experiments, and the bottom row is the bias spread experiments.
    We apply a 5-epoch sliding window in the bias spread experiment for a better visualization.
  }\label{fig:exp_large}
\end{figure*}
We first test the influence of each technique separately.
Comparing the performance of the biased and unbiased versions of PG in \cref{fig:exp_a}, we find the vanilla PG is harmed by the bias.
This echos back our discussions in \cref{sec:bias}.
Fortunately, all three techniques discussed in \cref{sec:rectification} show a strong bias rectification ability.

\paragraph{Learning Rate.}
\cref{fig:exp_a} shows that simply decaying the learning rate significantly mitigates the influence of bias. 
Under a smaller learning rate, the biased and unbiased versions of PG achieve very similar average performance with a smaller variance.
However, simply decaying the learning rate slows down the training procedure at the beginning.
Additionally, as the optimal learning rate varies for tasks, we turn to some adaptive learning rate methods.

\paragraph{Optimizer.}
Adaptive learning rate optimizers like RMSProp and Adam use empirical FIM to estimate the curvature of the loss surface to stabilize training.
In the experiment in \cref{fig:exp_b}, the RMSProp enhances the performance of PG and mitigates the influence of bias.
Adam can be viewed as a variant of RMSProp, with an additional momentum term.
It achieves a slightly better performance compared to RMSProp.
Using the Momentum optimizer alone without RMSProp hardly helps fix the bias.

\paragraph{Regularization.}
In \cref{fig:exp_c}, the performance and variance of three approaches are almost the same: the KL regularized unbiased version (unbiased+KL), the KL regularized biased version (biased+KL) and the reverse KL regularized biased version (biased+reverse KL).
These three approaches also have very small variances, consistent with our analysis in \cref{subsec:regu} that the regularization restricts the update in a small trust region.
Combining the RMSProp optimizer and KL regularization, we can achieve the best performance, with the smallest variance, high convergence speed and almost uninfluenced by the bias.

Additionally, we conduct a large-scale experiment on continuous control Mujoco benchmark~\cite{brockman2016openai} in \cref{fig:exp_large} to investigate the bias rectification ability of RMSProp optimizer combined with KL regularization.
We use \textit{correction} to stand for RMSProp+KL for simplicity.
The top row is the performance experiments to compare the unbiased and biased versions of PG with or without correction.
The bottom row is the bias spread experiments described in \cref{fig:bias_spread}. 
We compute the distance $\dis_1$ between the unbiased+correction version and the biased+correction version, and the distance $\dis_2$ between the unbiased version and the biased version.
A percentage distance $\dis \% = (\dis_1 - \dis_2) / (\dis_1+\dis_2) \in \left[-1, 1\right]$ is also computed to quantify the amount of bias-fixing of the correction method.
A $\dis \% < 0$ indicates ours is less influenced by the bias.
In the Inverted Pendulum and Reacher environments, the vanilla PG is severely harmed by the bias, but the correction method excellently fixes it.
In the Swimmer, Hopper and Inverted Double Pendulum environments, both the biased and unbiased versions of vanilla PG reach a local optimal quickly (ceiling effect) in the performance experiment.
From the bias spread experiment, we find that in all five environments, using RMSProp as the optimizer and adding the KL regularization term significantly reduce the bias.
The experiment details and the hyperparameter choices are provided in \cref{sec:hyper}.

\subsection{General Bias Rectification}

\begin{figure}[t]
  \centering
  \includegraphics[width=0.85\linewidth]{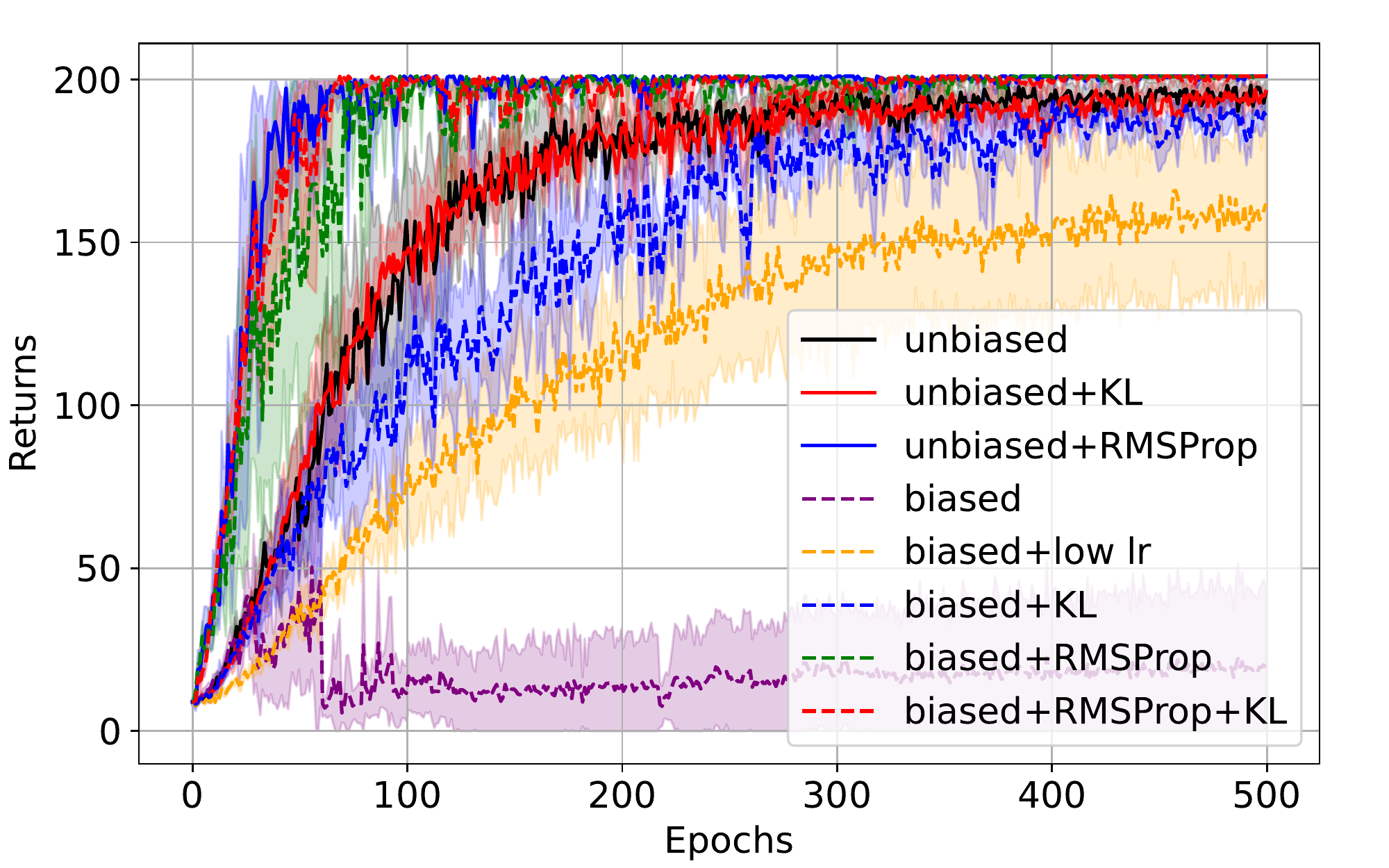}
  \caption{Off-policy bias rectification experiment (5 random seeds).}\label{fig:exp_off}
\end{figure}

The previous experiments suggest that the RMSProp optimizer and KL regularization might be helpful for a larger range of biases.
We simulate the off-policy bias by perturbing the state distribution described in \cref{appendix:general_bias}.
The result of the off-policy bias rectification performance experiment is shown in \cref{fig:exp_off}.
The biased PG collapses at the early stage without the help of any regularization or optimizer.
However, the aforementioned methods, low learning rate, regularization and RMSProp optimizer, still work well under this harsh perturbation.

In the future, extending our analysis to a more general state distribution shift such as in the off-policy setting is an interesting direction. 
We suggest these might be helpful for future off-policy PG algorithm design: 1) directly reducing the occurring possibility of the state alias phenomenon; 2) adopting some techniques like KL regularization to mitigate the influence of bias.
For 1), there are already some works on representation learning to distinguish aliased state, some are provided in \cref{sec:bias}; this paper further provides reasons to support the importance of representation learning.
For 2), we briefly discuss it in \cref{fig:exp_off}, but detailed theoretical study and experiments are left for future work.



\section{Conclusions and Future Work}
We study the long-concerned undiscounted bias in the policy gradient estimation in the DRL setting.
We introduce the state alias phenomenon, under which the biased policy gradient theoretically leads to sub-optimal behavior. 
Then, we find that some common but overlooked techniques, including reducing the learning rate, adaptive optimizers, and KL divergence regularization, can mitigate the bias effectively.
We support our findings with experiments in the Mujoco continuous control environments.

Avenues for future work include: 1) quantifying the occurring possibility of the state alias phenomenon; 2) developing algorithms mitigating the state alias to enhance the performance of PG under the off-policy setting; 3) further investigating the connections between optimizers and the optimization regularization.



\bibliographystyle{named}
\newpage
\bibliography{ijcai23}

\begin{thebibliography}{}

\bibitem[\protect\citeauthoryear{Abdolmaleki \bgroup \em et al.\egroup
  }{2018a}]{abdolmaleki2018relative}
Abbas Abdolmaleki, Jost~Tobias Springenberg, Jonas Degrave, Steven Bohez, Yuval
  Tassa, Dan Belov, Nicolas Heess, and Martin Riedmiller.
\newblock Relative entropy regularized policy iteration.
\newblock {\em arXiv preprint arXiv:1812.02256}, 2018.

\bibitem[\protect\citeauthoryear{Abdolmaleki \bgroup \em et al.\egroup
  }{2018b}]{abdolmaleki2018maximum}
Abbas Abdolmaleki, Jost~Tobias Springenberg, Yuval Tassa, Remi Munos, Nicolas
  Heess, and Martin Riedmiller.
\newblock Maximum a posteriori policy optimisation.
\newblock In {\em International Conference on Learning Representations}, 2018.

\bibitem[\protect\citeauthoryear{Agarwal \bgroup \em et al.\egroup
  }{2021a}]{agarwal2021reinforcement}
Alekh Agarwal, Nan Jiang, Sham~M Kakade, and Wen Sun.
\newblock Reinforcement learning: Theory and algorithms.
\newblock page 205, 2021.

\bibitem[\protect\citeauthoryear{Agarwal \bgroup \em et al.\egroup
  }{2021b}]{agarwal2021theory}
Alekh Agarwal, Sham~M Kakade, Jason~D Lee, and Gaurav Mahajan.
\newblock On the theory of policy gradient methods: Optimality, approximation,
  and distribution shift.
\newblock {\em J. Mach. Learn. Res.}, 22(98):1--76, 2021.

\bibitem[\protect\citeauthoryear{Agarwal \bgroup \em et al.\egroup
  }{2021c}]{agarwal2021contrastive}
Rishabh Agarwal, Marlos~C. Machado, Pablo~Samuel Castro, and Marc~G Bellemare.
\newblock Contrastive behavioral similarity embeddings for generalization in
  reinforcement learning.
\newblock In {\em International Conference on Learning Representations}, 2021.

\bibitem[\protect\citeauthoryear{Amari}{2012}]{amari2012differential}
Shun-ichi Amari.
\newblock {\em Differential-geometrical methods in statistics}, volume~28.
\newblock Springer Science \& Business Media, 2012.

\bibitem[\protect\citeauthoryear{Andrychowicz \bgroup \em et al.\egroup
  }{2021}]{andrychowicz2021what}
Marcin Andrychowicz, Anton Raichuk, Piotr Sta{\'n}czyk, Manu Orsini, Sertan
  Girgin, Rapha{\"e}l Marinier, Leonard Hussenot, Matthieu Geist, Olivier
  Pietquin, Marcin Michalski, Sylvain Gelly, and Olivier Bachem.
\newblock What matters for on-policy deep actor-critic methods? {A} large-scale
  study.
\newblock In {\em International Conference on Learning Representations}, 2021.

\bibitem[\protect\citeauthoryear{Becker and Lecun}{1989}]{becker1988improving}
Suzanna Becker and Yann Lecun.
\newblock Improving the convergence of back-propagation learning with
  second-order methods.
\newblock January 1989.

\bibitem[\protect\citeauthoryear{Brandfonbrener \bgroup \em et al.\egroup
  }{2021}]{brandfonbrener2021offline}
David Brandfonbrener, Will Whitney, Rajesh Ranganath, and Joan Bruna.
\newblock Offline {RL} without off-policy evaluation.
\newblock In {\em Advances in Neural Information Processing Systems},
  volume~34, pages 4933--4946, 2021.

\bibitem[\protect\citeauthoryear{Brockman \bgroup \em et al.\egroup
  }{2016}]{brockman2016openai}
Greg Brockman, Vicki Cheung, Ludwig Pettersson, Jonas Schneider, John Schulman,
  Jie Tang, and Wojciech Zaremba.
\newblock {OpenAI} gym, June 2016.

\bibitem[\protect\citeauthoryear{Cobbe \bgroup \em et al.\egroup
  }{2021}]{cobbe2021phasic}
Karl~W Cobbe, Jacob Hilton, Oleg Klimov, and John Schulman.
\newblock Phasic policy gradient.
\newblock In {\em International Conference on Machine Learning}, pages
  2020--2027. PMLR, 2021.

\bibitem[\protect\citeauthoryear{Duchi \bgroup \em et al.\egroup
  }{2011}]{duchi2011adaptive}
John Duchi, Elad Hazan, and Yoram Singer.
\newblock Adaptive subgradient methods for online learning and stochastic
  optimization.
\newblock {\em J. Mach. Learn. Res.}, 12(7), 2011.

\bibitem[\protect\citeauthoryear{Engstrom \bgroup \em et al.\egroup
  }{2019}]{engstrom2019implementation}
Logan Engstrom, Andrew Ilyas, Shibani Santurkar, Dimitris Tsipras, Firdaus
  Janoos, Larry Rudolph, and Aleksander Madry.
\newblock Implementation matters in deep {RL}: A case study on {PPO} and
  {TRPO}.
\newblock In {\em International conference on learning representations}, 2019.

\bibitem[\protect\citeauthoryear{Fakoor \bgroup \em et al.\egroup
  }{2020}]{fakoor2020p3o}
Rasool Fakoor, Pratik Chaudhari, and Alexander~J Smola.
\newblock P3o: Policy-on policy-off policy optimization.
\newblock In {\em Uncertainty in Artificial Intelligence}, pages 1017--1027.
  PMLR, 2020.

\bibitem[\protect\citeauthoryear{Frazier and
  Riedl}{2019}]{frazier2019improving}
Spencer Frazier and Mark Riedl.
\newblock Improving deep reinforcement learning in minecraft with action
  advice.
\newblock In {\em Proceedings of the AAAI conference on artificial intelligence
  and interactive digital entertainment}, volume~15, pages 146--152, 2019.

\bibitem[\protect\citeauthoryear{Fujimoto \bgroup \em et al.\egroup
  }{2018}]{fujimoto2018addressing}
Scott Fujimoto, Herke Hoof, and David Meger.
\newblock Addressing function approximation error in actor-critic methods.
\newblock In {\em International conference on machine learning}, pages
  1587--1596. PMLR, 2018.

\bibitem[\protect\citeauthoryear{Goodfellow \bgroup \em et al.\egroup
  }{2016}]{goodfellow2016deep}
Ian~J. Goodfellow, Yoshua Bengio, and Aaron Courville.
\newblock {\em Deep Learning}.
\newblock MIT Press, 2016.

\bibitem[\protect\citeauthoryear{Hessel \bgroup \em et al.\egroup
  }{2021}]{hessel2021muesli}
Matteo Hessel, Ivo Danihelka, Fabio Viola, Arthur Guez, Simon Schmitt, Laurent
  Sifre, Theophane Weber, David Silver, and Hado Van~Hasselt.
\newblock Muesli: Combining improvements in policy optimization.
\newblock In {\em International Conference on Machine Learning}, pages
  4214--4226. PMLR, 2021.

\bibitem[\protect\citeauthoryear{Ho and Ermon}{2016}]{ho2016generative}
Jonathan Ho and Stefano Ermon.
\newblock Generative adversarial imitation learning.
\newblock In {\em Advances in Neural Information Processing Systems},
  volume~29. Curran Associates, Inc., 2016.

\bibitem[\protect\citeauthoryear{Jacob \bgroup \em et al.\egroup
  }{2022}]{jacob2022modeling}
Athul~Paul Jacob, David~J Wu, Gabriele Farina, Adam Lerer, Hengyuan Hu, Anton
  Bakhtin, Jacob Andreas, and Noam Brown.
\newblock Modeling strong and human-like gameplay with {KL}-regularized search.
\newblock In {\em International Conference on Machine Learning}, pages
  9695--9728. PMLR, 2022.

\bibitem[\protect\citeauthoryear{Kingma and Ba}{2014}]{kingma2014adam}
Diederik~P Kingma and Jimmy Ba.
\newblock Adam: {A} method for stochastic optimization.
\newblock {\em arXiv preprint arXiv:1412.6980}, 2014.

\bibitem[\protect\citeauthoryear{Kuba \bgroup \em et al.\egroup
  }{2022}]{kuba2022trust}
Jakub~Grudzien Kuba, Ruiqing Chen, Muning Wen, Ying Wen, Fanglei Sun, Jun Wang,
  and Yaodong Yang.
\newblock Trust region policy optimisation in multi-agent reinforcement
  learning.
\newblock In {\em International Conference on Learning Representations}, 2022.

\bibitem[\protect\citeauthoryear{Laroche and Tachet~des
  Combes}{2021}]{laroche2021dr}
Romain Laroche and Remi Tachet~des Combes.
\newblock Dr jekyll \& mr hyde: The strange case of off-policy policy updates.
\newblock 34:24442--24454, 2021.

\bibitem[\protect\citeauthoryear{Lazi{\'c} \bgroup \em et al.\egroup
  }{2021}]{lazic2021optimization}
Nevena Lazi{\'c}, Botao Hao, Yasin Abbasi-Yadkori, Dale Schuurmans, and Csaba
  Szepesv{\'a}ri.
\newblock Optimization issues in {KL}-constrained approximate policy iteration.
\newblock {\em arXiv preprint arXiv:2102.06234}, 2021.

\bibitem[\protect\citeauthoryear{Lehnert and
  Littman}{2020}]{lehnert2020successor}
Lucas Lehnert and Michael~L Littman.
\newblock Successor features combine elements of model-free and model-based
  reinforcement learning.
\newblock {\em J. Mach. Learn. Res.}, 21:196--1, 2020.

\bibitem[\protect\citeauthoryear{Li \bgroup \em et al.\egroup
  }{2018}]{li2018visualizing}
Hao Li, Zheng Xu, Gavin Taylor, Christoph Studer, and Tom Goldstein.
\newblock Visualizing the loss landscape of neural nets.
\newblock In {\em Advances in Neural Information Processing Systems},
  volume~31. Curran Associates, Inc., 2018.

\bibitem[\protect\citeauthoryear{Li \bgroup \em et al.\egroup
  }{2021}]{li2021fault}
Yi~Li, Shaohua Wang, and Tien Nguyen.
\newblock Fault localization with code coverage representation learning.
\newblock In {\em IEEE/ACM International Conference on Software Engineering},
  pages 661--673. IEEE, 2021.

\bibitem[\protect\citeauthoryear{Liu \bgroup \em et al.\egroup
  }{2021}]{liu2021returnbased}
Guoqing Liu, Chuheng Zhang, Li~Zhao, Tao Qin, Jinhua Zhu, Li~Jian, Nenghai Yu,
  and Tie-Yan Liu.
\newblock Return-based contrastive representation learning for reinforcement
  learning.
\newblock In {\em International Conference on Learning Representations}, 2021.

\bibitem[\protect\citeauthoryear{Nachum \bgroup \em et al.\egroup
  }{2018}]{nachum2018trustpcl}
Ofir Nachum, Mohammad Norouzi, Kelvin Xu, and Dale Schuurmans.
\newblock Trust-{PCL}: An off-policy trust region method for continuous
  control.
\newblock In {\em International Conference on Learning Representations}, 2018.

\bibitem[\protect\citeauthoryear{Nota and Thomas}{2020}]{nota2020policy}
Chris Nota and Philip~S. Thomas.
\newblock Is the policy gradient a gradient?
\newblock In {\em International Conference on Autonomous Agents and MultiAgent
  Systems}, page 939–947, 2020.

\bibitem[\protect\citeauthoryear{Reddi \bgroup \em et al.\egroup
  }{2018}]{reddi2019convergence}
Sashank~J. Reddi, Satyen Kale, and Sanjiv Kumar.
\newblock On the convergence of {Adam} and beyond.
\newblock In {\em International Conference on Learning Representations}, 2018.

\bibitem[\protect\citeauthoryear{Schulman \bgroup \em et al.\egroup
  }{2015}]{schulman2015trust}
John Schulman, Sergey Levine, Pieter Abbeel, Michael Jordan, and Philipp
  Moritz.
\newblock Trust region policy optimization.
\newblock In {\em International Conference on Machine Learning}, pages
  1889--1897. PMLR, June 2015.

\bibitem[\protect\citeauthoryear{Schulman \bgroup \em et al.\egroup
  }{2017}]{schulman2017proximal}
John Schulman, Filip Wolski, Prafulla Dhariwal, Alec Radford, and Oleg Klimov.
\newblock Proximal policy optimization algorithms, August 2017.

\bibitem[\protect\citeauthoryear{Shani \bgroup \em et al.\egroup
  }{2020}]{shani2020adaptive}
Lior Shani, Yonathan Efroni, and Shie Mannor.
\newblock Adaptive trust region policy optimization: Global convergence and
  faster rates for regularized {MDPs}.
\newblock In {\em Proceedings of the AAAI Conference on Artificial
  Intelligence}, volume~34, pages 5668--5675, 2020.

\bibitem[\protect\citeauthoryear{Silver \bgroup \em et al.\egroup
  }{2014}]{silver2014deterministic}
David Silver, Guy Lever, Nicolas Heess, Thomas Degris, Daan Wierstra, and
  Martin Riedmiller.
\newblock Deterministic policy gradient algorithms.
\newblock In {\em International Conference on Machine Learning}, pages
  387--395. PMLR, January 2014.

\bibitem[\protect\citeauthoryear{Stooke and Abbeel}{2019}]{stooke2019rlpyt}
Adam Stooke and Pieter Abbeel.
\newblock rlpyt: A research code base for deep reinforcement learning in
  pytorch.
\newblock {\em arXiv preprint arXiv:1909.01500}, 2019.

\bibitem[\protect\citeauthoryear{Sun \bgroup \em et al.\egroup
  }{2022}]{sun2022you}
Mingfei Sun, Vitaly Kurin, Guoqing Liu, Sam Devlin, Tao Qin, Katja Hofmann, and
  Shimon Whiteson.
\newblock You may not need ratio clipping in ppo.
\newblock {\em arXiv preprint arXiv:2202.00079}, 2022.

\bibitem[\protect\citeauthoryear{Sutton and
  Barto}{2018}]{sutton2018reinforcement}
Richard~S Sutton and Andrew~G Barto.
\newblock {\em Reinforcement learning: An introduction}.
\newblock MIT press, 2018.

\bibitem[\protect\citeauthoryear{Sutton \bgroup \em et al.\egroup
  }{1999}]{sutton1999policy}
Richard~S Sutton, David McAllester, Satinder Singh, and Yishay Mansour.
\newblock Policy gradient methods for reinforcement learning with function
  approximation.
\newblock In {\em Advances in Neural Information Processing Systems},
  volume~12. MIT Press, 1999.

\bibitem[\protect\citeauthoryear{Thomas}{2014}]{thomas2014bias}
Philip~S Thomas.
\newblock Bias in natural actor-critic algorithms.
\newblock In {\em International Conference on Machine Learning}, page~8, 2014.

\bibitem[\protect\citeauthoryear{Tieleman and
  Hinton}{2012}]{Tieleman2012RMSProp}
T.~Tieleman and G.~Hinton.
\newblock Lecture 6.5 - {RMSProp}: Divide the gradient by a running average of
  its recent magnitude.
\newblock {\em Course notes for CS 294: Deep Learning}, 2012.

\bibitem[\protect\citeauthoryear{Todorov \bgroup \em et al.\egroup
  }{2012}]{todorov2012mujoco}
Emanuel Todorov, Tom Erez, and Yuval Tassa.
\newblock Mujoco: A physics engine for model-based control.
\newblock In {\em 2012 IEEE/RSJ International Conference on Intelligent Robots
  and Systems}, pages 5026--5033. IEEE, 2012.

\bibitem[\protect\citeauthoryear{Uehara \bgroup \em et al.\egroup
  }{2022}]{uehara2022representation}
Masatoshi Uehara, Xuezhou Zhang, and Wen Sun.
\newblock Representation learning for online and offline {RL} in low-rank
  {MDP}s.
\newblock In {\em International Conference on Learning Representations}, 2022.

\bibitem[\protect\citeauthoryear{Vieillard \bgroup \em et al.\egroup
  }{2020}]{vieillard2020leverage}
Nino Vieillard, Tadashi Kozuno, Bruno Scherrer, Olivier Pietquin, R{\'e}mi
  Munos, and Matthieu Geist.
\newblock Leverage the average: {An} analysis of {KL} regularization in
  reinforcement learning.
\newblock 33:12163--12174, 2020.

\bibitem[\protect\citeauthoryear{Wu \bgroup \em et al.\egroup
  }{2020}]{wu2020reducing}
Dongming Wu, Xingping Dong, Jianbing Shen, and Steven~CH Hoi.
\newblock Reducing estimation bias via triplet-average deep deterministic
  policy gradient.
\newblock {\em IEEE transactions on neural networks and learning systems},
  31(11):4933--4945, 2020.

\bibitem[\protect\citeauthoryear{Wu \bgroup \em et al.\egroup
  }{2022}]{wu2022understanding}
Shuang Wu, Ling Shi, Jun Wang, and Guangjian Tian.
\newblock Understanding policy gradient algorithms: A sensitivity-based
  approach.
\newblock In {\em International Conference on Machine Learning}, page~19, 2022.

\bibitem[\protect\citeauthoryear{Zeiler}{2012}]{zeiler2012adadelta}
Matthew~D Zeiler.
\newblock Adadelta: {An} adaptive learning rate method.
\newblock {\em arXiv preprint arXiv:1212.5701}, 2012.

\bibitem[\protect\citeauthoryear{Zhan \bgroup \em et al.\egroup
  }{2021}]{zhan2021policy}
Wenhao Zhan, Shicong Cen, Baihe Huang, Yuxin Chen, Jason~D Lee, and Yuejie Chi.
\newblock Policy mirror descent for regularized reinforcement learning: A
  generalized framework with linear convergence.
\newblock {\em arXiv preprint arXiv:2105.11066}, 2021.

\bibitem[\protect\citeauthoryear{Zinkevich}{2003}]{zinkevich2003online}
Martin Zinkevich.
\newblock Online convex programming and generalized infinitesimal gradient
  ascent.
\newblock In {\em International Conference on Machine Learning}, pages
  928--936, 2003.

\bibitem[\protect\citeauthoryear{Zou \bgroup \em et al.\egroup
  }{2019}]{zou2019sufficienta}
Fangyu Zou, Li~Shen, Zequn Jie, Weizhong Zhang, and Wei Liu.
\newblock A sufficient condition for convergences of {Adam} and {RMSProp}.
\newblock In {\em IEEE/CVF Conference on Computer Vision and Pattern
  Recognition}, pages 11119--11127, Long Beach, CA, USA, June 2019.

\end{thebibliography}


\newpage
\appendix
\onecolumn
\section{Degenerated Example for \Cref{fig:alias}}\label{appendix: alias}
\begin{figure}[ht]
  \centering
  \includegraphics[width=0.6\linewidth]{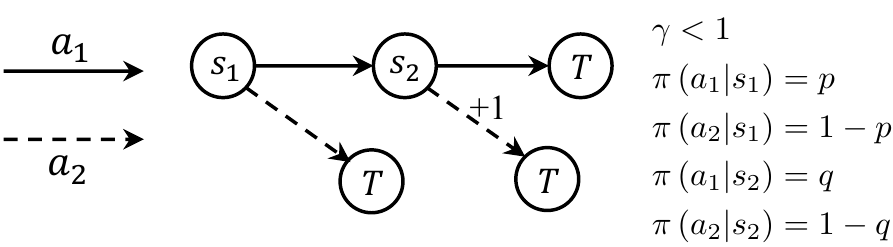}
  \caption{
    Same as \Cref{fig:alias}. An example to clarify why the discount factor in state distribution is important. 
  }
  \label{fig:appendix_1}
\end{figure}
The expected discounted return of this example is
\[
  \rho(p, q) = p (1-q) \gamma.
\]
We describe the state alias phenomenon with the following constraint:
\begin{equation*}
  \max_a \| \pi(a|s_1) - \pi(a|s_2) \| \leq \epsilon, \epsilon < \gamma.
\end{equation*}
When $\epsilon = 0$, the constraint degenerates to
\[p=q.\]
Easily we can compute the optimal policy:
$$
  p^* = q^* = \argmax_{p,q} \rho(p, q) = \frac{1}{2}
$$
Now we simulate the iterative policy optimization process. 

With a similar deduction in \cref{sec:bias}, we can compute the converged policies
\begin{align}
  p_\gamma^* &= \frac{1}{2} 
  &q_\gamma^* &= \frac{1}{2} 
  &\rho(p_\gamma^*, q_\gamma^*) &= \frac{1}{4}\gamma
  \\
  p_1^* &= \frac{\gamma}{1+\gamma} 
  &q_1^* &= \frac{\gamma}{1+\gamma}  
  &\rho(p_1^*, q_1^*) &= \frac{\gamma^2}{(1+\gamma)^2}.
\end{align}
The performance decay of biased PG is
\[\frac{\rho(p_1^*)}{\rho(p_\gamma^*)} = \frac{4\gamma}{(1+\gamma)^2} < 1.\]
We note that when $\gamma$ approaches $1$, the bias would reduce, which echos back the findings in \cite{wu2022understanding}.
A similar phenomenon was first introduced by~\cite{nota2020policy}.

\section{Deivation of the Reverse KL Regularized Objective Function}\label{appendix: reverse_kl}
We provide the derivation of the reverse KL regularized objective function
\begin{align}
  J(\pi|\pi_t) &=\mathbb{E}_{s\sim d_{\pi_t}}\mathbb E_{a\sim \pi_t}\left[{q}_{\pi_t} \right]-\beta K L\left(\pi, \pi_t\right)\\
  &=\mathbb{E}_{s\sim d_{\pi_t}}\mathbb E_{a\sim \pi_t}\left[\pi q_{\pi_t}\right] +\beta\pi \log\pi_t -\beta \pi \log\pi \\
  &= \mathbb{E}_{s\sim d_{\pi_t}}\mathbb E_{a\sim \pi_t}\left[\frac{\pi}{\pi_t}q_{\pi_t}+\beta\frac{\pi}{\pi_t}\left(\log\pi_t-\log\pi\right)\right].
\end{align}

\section{Complete Analysis of the State Alias in DRL} \label{appendix:c}
We expand the discussions in \cref{sec:bias} and \cref{fig:alias_nn} here.
We aim to examine how the feature representations of states evolve during the training procedure.
We first train a successful model $\pi^*$ converged to the optimal policy in the Inverted Pendulum environment (with an average return=$195$).
We collect a huge amount of data $\mathcal{D}=\{(s, a)\}$ containing the state and action pairs with a mid-way model separately (not used along training).
The mid-way model is a model which could reach the max return ($200$) but with the average return=$100$.
We use this model because it can reach both the data with high expected return and low expected return.
Then, we feed those states into different immature models along the training procedure and extract the feature representations.
We use $f^i_\pi(s)$ to denote the feature representations of the state $s$ passing the first $i$ layers of the neural network of the model $\pi$.
We compare the feature representations of immature models with the optimal model.
Then we visualize the feature representations in a 2D surface with the help of PCA and mark the action value with different colors.
We also compute a correlation coefficient of the projected feature representation value (x and y axis) and action value.
Here we use $\pi^{x\%}$ to denote the model with a $x\%$ average return of the optimal model.

We report the result of comparing the optimal model representations $\{(f^2_{\pi^*}(s), a)\}$ with the early stage model representations $\{(f^2_{\pi^{10\%}}(s), a)\}$ in \cref{fig:alias_nn}.

\begin{figure*}[ht]
  \begin{minipage}[t]{0.24\linewidth}
    \includegraphics[width=0.95\linewidth]{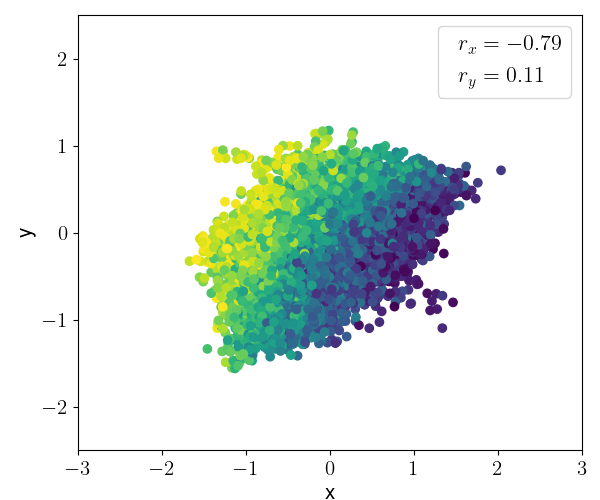}
    \includegraphics[width=0.95\linewidth]{sections/images/similar_state/layer_2_return_20.png}
    \subcaption{$\pi^{10\%}$}
  \end{minipage}
  \begin{minipage}[t]{0.24\linewidth}
      \includegraphics[width=0.95\linewidth]{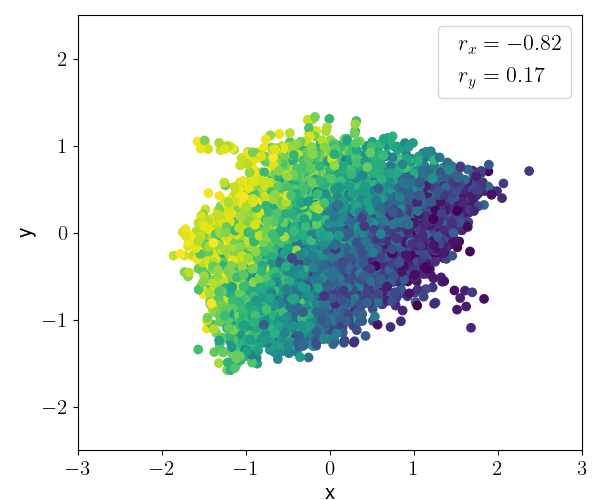}
      \includegraphics[width=0.95\linewidth]{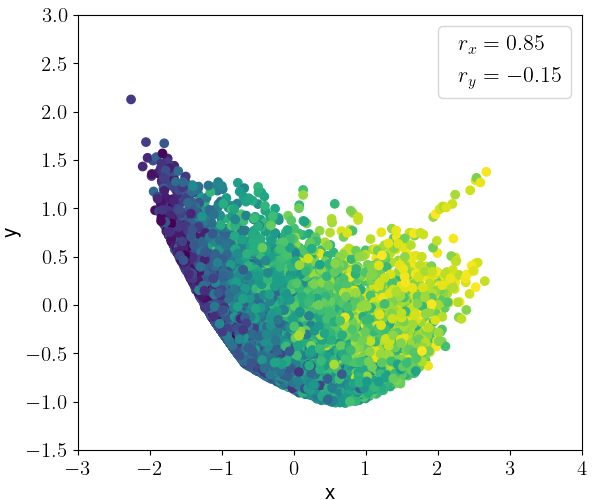}
      \subcaption{$\pi^{25\%}$}
  \end{minipage}
  \begin{minipage}[t]{0.24\linewidth}
    \includegraphics[width=0.95\linewidth]{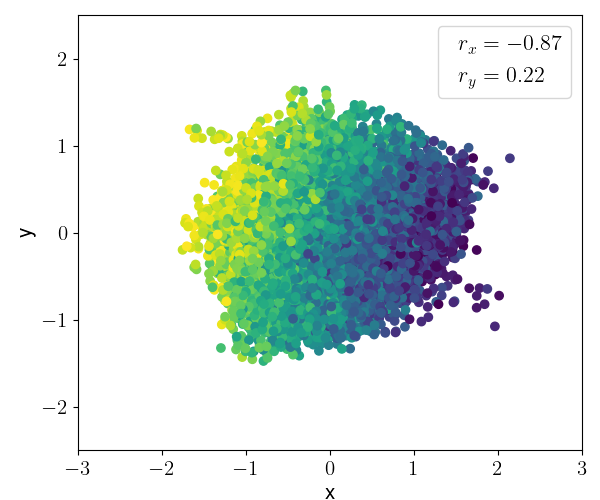}
      \includegraphics[width=0.95\linewidth]{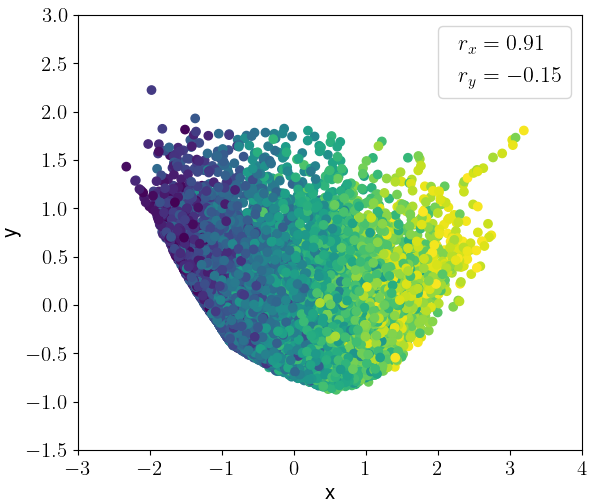}
      \subcaption{$\pi^{50\%}$}
  \end{minipage}
  \begin{minipage}[t]{0.24\linewidth}
      \includegraphics[width=0.95\linewidth]{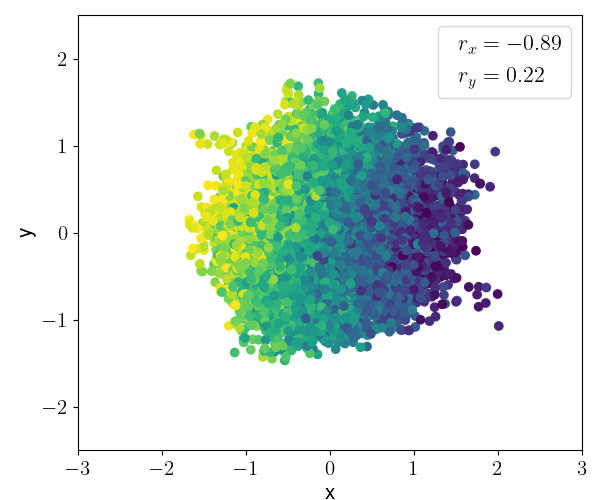}
      \includegraphics[width=0.95\linewidth]{sections/images/similar_state/layer_2_return_190.png}
      \subcaption{$\pi^*$}
  \end{minipage}
  \caption{We feed the same state dataset to different models representing different stages of the training procedure, extract the features from the first/second layer of the model, then map them to 2D space with PCA. 
  Left to right: 4 models with return=20, 50, 100, 190. 
  Up: the first layer; Down: the second layer.
  As in Inverted Pendulum, the dimension of action space is 1, we use the color to indicate the action.
  }\label{fig:appendix_alias_large}
\end{figure*}

A more detailed result with different layers and different stages are provided in \cref{fig:appendix_alias_large}.

\section{Experiments Details}
\subsection{Network Stucture}
We use a small network with 2 hidden layers, 16 units per layer and a ReLU activation. 
Then the output features are sent into an actor layer to predict a policy distribution.

\subsection{Loss Surfaces Analysis in \cref{fig:loss_surface}} \label{appendix:d2}
This section provides some details of visualizing the loss surface in \cref{fig:loss_surface}.

We mainly follow \cite{li2018visualizing}'s filter normalization method and codes to visualize the loss surface.
Their original work is for supervised learning; we manage to adapt it to the reinforcement learning setting by introducing a "score model" to compute the score function.
We collect these data for the score model: state $s$, action $a$ and return $r$.
With a huge number of data (almost $10\times$ more than the data needed to train the reinforcement learning model) $\mathcal{D} = \{(s, a, r)\}$, we separately train a supervised model $\mathcal{T}$ to predict return $r$ given state $s$ and action $a$.
Then we can draw the loss surface by computing the loss for the unregularized form given by
\begin{equation}
  \mathcal{L}(\theta) =  -\mathbb{E}_{s \sim \mathcal{D}}\mathbb{E}_{a \sim \pi_\theta}\left[\mathcal{T}(s,a)\right].
\end{equation}
The loss for the KL regularized form is given by
\begin{equation}
  \mathcal{L}(\theta) =  -\mathbb{E}_{s \sim \mathcal{D}}\mathbb{E}_{a \sim \pi_\theta}\left[\mathcal{T}(s,a)+\log(\pi_\theta(a|s))\right].
\end{equation}
Then, we follow the author's method to plot the loss surface. Please refer to  \cite{li2018visualizing} for a detailed description.

\subsection{Hyperparameter for Mujoco Environments}\label{sec:hyper}
We truncate the episode at 200 steps for all the experiments.
After collecting $N$ episodes of data, we receive a dataset $\mathcal{D}$ with size $|\mathcal D|$. 
The basic learning rate $lr$ is set at the beginning of the training procedure and exponentially decayed by a factor of $d$ every $e$ epochs.
We train the model with a learning rate $lr/|D| \times 1000$ by sampling data from $\mathcal D$ for $|\mathcal D|$ times.
We set the basic learning rate $lr$ and $N$ based on the following table.
\begin{table}
  \center
  \begin{tabular}{||c||c |c| c| c| c||} 
    \hline
    &Inverted Pendulum & Reacher & Swimmer & Hopper & Inverted Double Pendulum \\ 
    \hline
    $lr$&3e-4 & 1e-4 & 1e-4 & 1e-4 & 5e-5\\ 
    \hline
    $N$& 10 & 20 & 5 & 5 & 20\\ 
    \hline
    $d$& 0.8 & 0.9 & 0.95 & 0.95 & 0.95\\ 
    \hline
    $e$& 30 & 50 & 30 & 30 & 30\\ 
    \hline
  \end{tabular}
  \caption{Hyperparameters}
\end{table}
When adopting the KL regularized form \cref{eq:kl} and the reverse KL regularized form \cref{eq:reverse_kl}, we set $\alpha=0.3$ for all the experiments.
Except for the learning rate $lr$, We use the default hyperparameters in Pytorch for RMSProp and Adam optimizers.
In the low learning rate setting in the Inverted Pendulum environment, the $lr$ is set to be 1.5e-4 (half the origin).

\subsection{General Bias Rectification}\label{appendix:general_bias}
We simulate the off-policy state distribution by manually enlarging the sampling probability of some states.
In \cref{fig:exp_large}, we adopt the following strategy to disturb the state distribution:
when $abs(s[0]) < 0.01$, we enlarge the sampling probability of this state by 5 times.
We use the same hyperparameter in \cref{sec:hyper}, except for $\alpha=0.5$ for the KL regularization coefficient.

\end{document}